\documentclass[10pt,twocolumn,letterpaper]{article}

\usepackage{cvpr}              

\usepackage{pifont}
\newcommand{\cmark}{\ding{51}}%
\newcommand{\xmark}{\ding{55}}%

\usepackage{makecell}
\usepackage[accsupp]{axessibility}  

\definecolor{cvprblue}{rgb}{0.21,0.49,0.74}
\usepackage[pagebackref,breaklinks,colorlinks,allcolors=cvprblue]{hyperref}

\newcommand*{\dif}[1]{{#1}}

\def\paperID{11815} 
\def\confName{CVPR}
\def\confYear{2025}

\title{ProbeSDF: Light Field Probes For Neural Surface Reconstruction}

\author{Briac Toussaint$^{1}$\\
{\tt\small briac.toussaint@inria.fr}
\and
Diego Thomas$^{2}$\\
{\tt\small thomas@ait.kyushu-u.ac.jp}
\and
Jean-Sébastien Franco$^{1}$\\
{\tt\small jean-sebastien.franco@inria.fr}\\
$^{1}$Univ. Grenoble Alpes, CNRS, Inria, Grenoble INP, LJK, France\\
$^{2}$Kyushu University, Japan\\
}

\begin{document}
\maketitle
\begin{figure*}[h]
    \centering
    \includegraphics[width=0.9\linewidth]{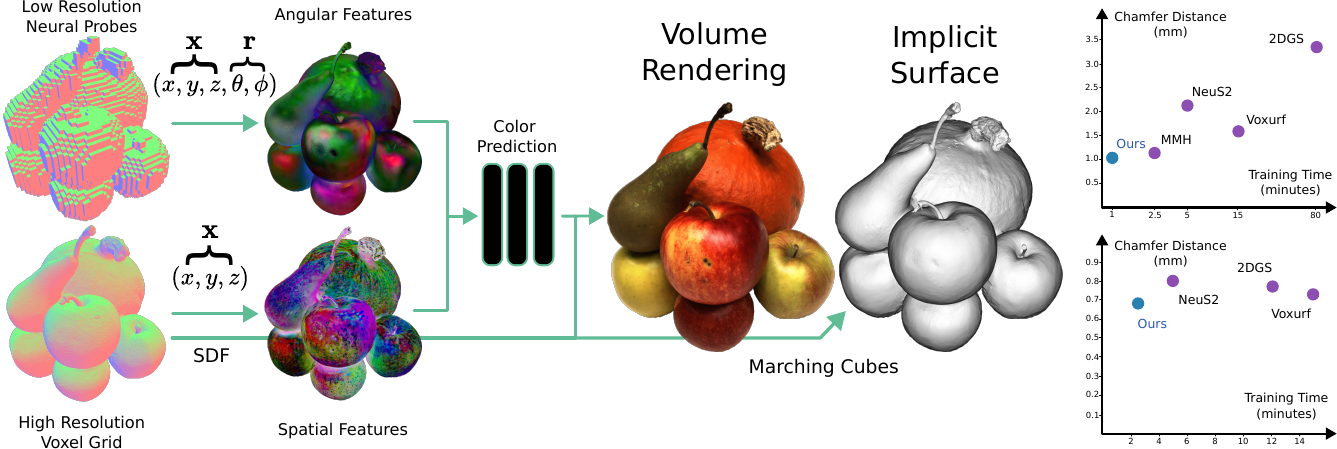}
    \caption{We design a new appearance model for neural surface approaches, which combines \dif{high resolution} spatial \dif{features} and \dif{lower resolution} angular features for improved reconstruction quality, training and inference speed. We plot the chamfer distance as a function of training speed for several baselines on MVMannequins (top) and DTU (bottom).}
    \label{fig:Teaser}
\end{figure*}
\begin{abstract}

SDF-based differential rendering frameworks have achieved state-of-the-art multiview 3D shape reconstruction. In this work, we re-examine this family of approaches by minimally reformulating its core appearance model in a way that simultaneously yields faster computation and increased performance. To this goal, we exhibit a physically-inspired minimal radiance parametrization decoupling angular and spatial contributions, by encoding them with a small number of features stored in two respective volumetric grids of different resolutions. Requiring as little as four parameters per voxel, and a tiny MLP call inside a single fully fused kernel, our approach allows to enhance performance with both surface and image (PSNR) metrics, while providing a significant training speedup and real-time rendering. We show this performance to be consistently achieved on real data over two widely different and popular application fields, generic object and human subject shape reconstruction, using four representative and challenging datasets.\footnote{\dif{Source code is available at \url{https://gitlab.inria.fr/projects-morpheo/ProbeSDF}}}
\end{abstract}

\section{Introduction}
\label{sec:intro}

Neural radiance fields have established a new milestone for the task of novel view synthesis and have led to a plethora of variants of broad applicability. We here take particular interest in SDF surface-based radiance fields, which have proven to be a highly performing variant for the tasks of 3D modeling from images~\cite{wang2021neus,wu2022voxurf}, with wide reaching applications ranging from object \dif{and} performance capture, to immersive and 3D content production. 

At the heart of such methods usually lies a  parametrization of the light field, in the form of a 5-dimensional function. This function maps the spatial coordinates of a point in space and the observed direction to a color and a signed distance to the reconstructed surface. On one hand, the mapping can be global and use a complex scene-wide MLP as implicit 5D decoder, as is the case for most seminal approaches~\cite{mildenhall2020nerf,wang2021neus}. On the other hand, recent popular variants that target higher speed and memory efficiency use explicit representations. Global representations are based on a grid encoding~\cite{wu2022voxurf,yu2022plenoxels,neus2} where interpolated spatial and angular features are run through a global shallow MLP. Local representations use the sparse 3D Gaussian Splatting~\cite{kerbl3Dgaussians} sample-based encoding approach, where each separate Gaussian comes with its own opacity and adjoining set of angular color features. While grids or local approaches use a single deep MLP to densely encode the complete light field function, they almost always co-locate storage and decoding of color and angular features. 

But is this really a necessary or even desirable feature of any light field encoding? We question these assertions by postulating that, contrary to the spatial components of the light field that mainly encode intrinsic surface texture, the angular components encode phenomena that are predominantly extrinsic to the surface, \textit{i.e.} lighting and environment, and as such can be encoded at a lower spatial resolution as they are locally  almost invariant to parallax. This hypothesis is in fact exploited by the rendering community, where angular and environment components contributing to the radiance equation used to compute rendered colors, are typically precomputed at sparse 3D locations coined \textit{light field probes}~\cite{ReadyAtDawn}, and interpolated in-between.

Leveraging this intuition, our approach borrows the spherical harmonic parametrization popular with explicit methods \eg~\cite{kerbl3Dgaussians,yu2022plenoxels} to represent angular  components, but instead decorrelates high density spatial variation from lower density angular variation using a mid-resolution probe grid of angular features. Each of the components encode abstract features that are fed to a tiny MLP, \dif{which} acts as a minimal BRDF decoder of spatial and angular varying components, while we explicitly store SDF values. This scheme advantageously replaces the core appearance and decoding model of SDF approaches \cite{wang2021neus,wu2022voxurf,neus2,li2023neuralangelo} to predict ray colors. We are careful to also include power terms of the cosine angle between the view direction and the estimated surface normal, allowing our lightweight model to effectively account for increased surface reflectivity at grazing angles. 
This yields an unprecedented performance combination, superseding both 3D and image metric performances with four popular benchmarks for general object reconstruction \cite{yao2020blendedmvs,jensen2014large} and human reconstruction \cite{millimetrichumans,isik2023humanrf} while simultaneously achieving a significant training and rendering performance boost.
In summary, our contributions are:
\begin{itemize}
    \item An explicit embedding of the angular dependency that is simultaneously local, smooth and efficient.
    \item A simpler color prediction MLP that is agnostic to the surface position and orientation.
\end{itemize}

\section{Related Works}
\label{sec:relworks}
In this work we leverage disentangled spatial and angular parameterization of radiance towards high quality appearance and surface modeling. Here we \dif{consequently} review closely related works on surface-based 3D scene representations and existing parametrizations of color.

\textbf{Neural Surface Reconstruction.}
Through the use of implicit surface representations, neural implicit fields have demonstrated increasing capabilities to generate high-quality 3D models \cite{niemeyer2020differentiable, yariv2020multiview, yariv2021volume}. In NeuS \cite{wang2021neus}, Wang et al. proposed to parametrize the surface with a signed distance function that is modeled with a multi layer perceptron (MLP). The approach opened new frontiers in the generation of accurate 3D scenes from images but at the cost of prohibitive training times (several hours). Follow-up works have thus focused on boosting both accuracy and computation time by employing explicit grids of features and smaller MLPs. Notably, Wu et al. proposed in Voxurf \cite{wu2022voxurf} to use a dense grid of voxels, and Neus2 was also proposed  \dif{using} hash grids \cite{neus2}. Recently, Millimetric humans \cite{millimetrichumans} have demonstrated remarkably fast and accurate reconstructions in the case of human datasets by using sparse and dynamic grids. \dif{Other reparametrizations target performance improvement, \eg  point-based with Fast Dipole Sums~\cite{Chen:Dipoles:2024} and PGSR~\cite{chen2024pgsr} both leveraging multi-view stereo priors, 
or by seeking mutual advantage of a hybrid volume/mesh representation~\cite{huang2024surf2f}. 
All the methods discussed above rely on co-located encodings of position and view direction to parametrize the appearance. As a consequence, these approaches either fail to model local radiance effects like inter-reflections, or require larger MLPs and  training times, and may thus benefit from our refactored appearance model as drop-in replacement.}

\textbf{Angular parametrization} Deep MLPs have become ubiquitous to model view-dependent colors, taking as input the view direction \cite{mildenhall2020nerf}, the view and normal \cite{yariv2020multiview, wang2021neus, wu2022voxurf, neus2} or the reflected direction \cite{verbin2022refnerf, millimetrichumans}. Non-neural approaches such as plenoxels \cite{yu2022plenoxels} or the Gaussian splatting methods \cite{kerbl3Dgaussians, Huang2DGS2024} favor a \dif{straightforward} spherical harmonics decomposition \dif{in color space}. The main advantage compared to a deep MLP is the computational efficiency, at the cost of a large number of appearance parameters: 27 per voxel for \cite{yu2022plenoxels} and 48 per Gaussian for \cite{kerbl3Dgaussians}. Note that some neural architectures \cite{verbin2022refnerf, neus2, millimetrichumans} use spherical harmonics as a positional encoding of the directional vectors, but this is completely distinct from encoding the signal into spherical harmonic coefficients. In this work we propose a lighter yet efficient parametrization of color by using two decorrelated voxel structures to encode different components of the observed color, \dif{and spherical harmonics encoding of \textit{pre-MLP features} for more parsimonious and expressive color decoding.}

In the field of real-time rendering, where lighting and BRDF are known in advance, the idea of caching the incoming light at discrete locations, known as light field probes, has become standard~\cite{majercik2020scaling, mcguire2017real, ReadyAtDawn}. We draw inspiration from this technique for our proposed approach to multi-view 3D reconstruction, with the difference in our case that the probe parameters are optimized along with the voxel parameters.

\textbf{High Frequency Reflections.} 
A number of approaches specifically deal with high frequency reflections, which requires dedicated representations and algorithms. Guo et al.~\cite{Guo_2022_CVPR} \dif{train an additional NeRF that is composited with the main image}. This approach is under-constrained and thus requires strong priors such as the reflective surfaces being flat. Verbin et al. \cite{verbin2022refnerf} \dif{model lighting as a neural envmap, modulated by a per-point roughness.} 
\dif{The roughness and base color of the object can be edited afterwards but a long training time is required due to the large MLP.}

Recent works \cite{verbin2024nerf, wu2024neural} recover detailed reflections using cone tracing, which inherently solves the problem posed by self-reflections. 
Both approaches give impressive results even when confronted with curved, mirror-like objects but at a large training cost exceeding 10 GPU-hours per scene. 
In contrast, we target objects with mild specularities,
and show that both fast training and real-time rendering can be achieved. More specifically, our neural light field probes can learn information about the surroundings whereas cone tracing gathers it at a greater cost. This assumption proves to be valid for standard acquisition scenes, with the four datasets tested.

\textbf{Minimal Parametrization.} 
Explicitly disentangling the material parameters from the lighting is particularly challenging but also rewarding because it can yield a minimal parametrization of the appearance. 
Munkberg et al. \cite{Munkberg_2022_CVPR} show that the approximate rendering equation developed in~\cite{karis2013real} is sufficiently accurate to recover the geometry using a differentiable rasterizer, along with an approximate material decomposition. 
Jiang et al. \cite{jiang2023gaussianshader} applies a similar strategy but with the 3D gaussian splatting volume rendering. The NeRFactor pipeline \cite{NeRFactor} starts from a pre-trained NeRF that is distilled into lighting, albedo and neural BRDF parameters. Jin et al. proposed TensoIR \cite{Jin2023TensoIR}, \dif{which} achieves a more accurate decomposition by evaluating the rendering equation with importance sampling. An efficient tensor representation helps to accelerate the rendering of the secondary rays, but the complete training still takes five hours on a single GPU, with decent results available after half an hour of training.
Our proposed method converges in just a few minutes and we find that only 4 coefficients per voxel are sufficient in most cases. The dimensionality is equivalent to the albedo and roughness values of a physically-based material parametrization, but without the cost associated with the inversion of the rendering equation. 

\begin{figure*}[h]
    \centering
    \includegraphics[width=1.0\textwidth]{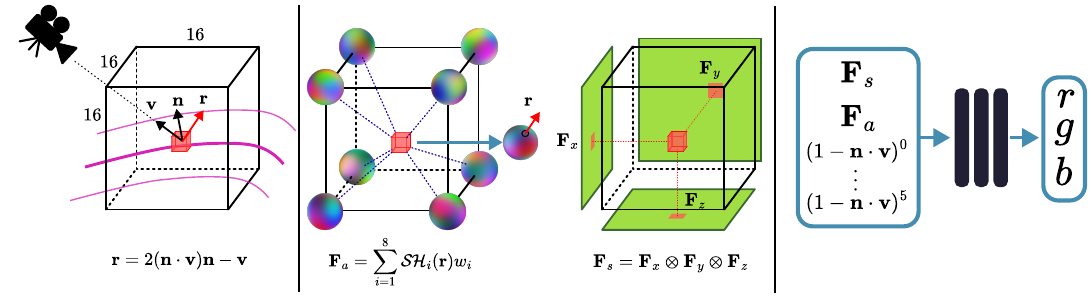}
    \caption{Overview of the color prediction for a single voxel inside a $16^3$ tile. The angular features $F_a$ are computed by interpolating and evaluating the spherical harmonics from the 8 \dif{corner probes} with the reflected vector $\mathbf{r}$. The spatial features $F_s$ are computed as the outer product of three orthogonal planes \dif{of resolution $16\times16$, specific to the tile}. A small neural network decodes these inputs into a color.}
    \label{fig:overview}
\end{figure*}

\section{Method}
\label{sec:method}

We use an SDF grid as the implicit representation of the 3D shape. SDF values in the grid are optimized via our newly proposed differentiable appearance model and the derived surface normal vectors $\mathbf{n}$.

\subsection{Decoupled Parametrization of Radiance}
\label{subsec:parametrization_preliminaries}

Volumetric neural rendering, as proposed in the seminal work NeRF, computes the color at a given pixel by integrating radiances along camera-pixel rays weighted by local opacity values. Follow-up papers condition the opacity on the distance to the surface to improve the reconstruction quality \cite{wang2021neus, yariv2021volume, Miller:VOS:2024}. The radiance $C$ represents the amount of energy that is emitted by a point in space. In physics it is expressed as the integral over the hemisphere $\Omega$ of the spatially varying bidirectional reflectance distribution function (SVBRDF) $f_r$ multiplied by the incoming radiance $L$ and by the cosine between the incoming direction $\mathbf{\omega}$ and the surface normal $\mathbf{n}$:
\begin{align}
    C &= \int_\Omega f_r(\mathbf{v}, \mathbf{\omega}, \mathbf{x}) L(\mathbf{\omega}, \mathbf{x}) (\mathbf{\omega} \cdot \mathbf{n}) d\mathbf{\omega}.
    \label{eq:rendering}
\end{align}
The (spatially varying) BRDF depends on the point's position $\mathbf{x}$, direction of incoming light $\mathbf{\omega}$ and viewing direction~$\mathbf{v}$. $L(\mathbf{\omega}, \mathbf{x})$ is the intensity of light at point $\mathbf{x}$ that comes from the direction $\mathbf{\omega}$ and $\mathbf{n}$ is the normal vector of the surface at position $\mathbf{x}$. In equation \ref{eq:rendering}, both $f_r$ and $L$ are unknown and disentangling these two functions during optimization is extremely difficult. As a consequence, \dif{a current research trend is to use} an MLP to directly approximate the \textit{result} of this integral and obtain the outgoing radiance with the following equation:
\begin{align}
    C &= MLP(\mathbf{F}_s(\mathbf{x}), \mathbf{v}, \mathbf{n}, \mathbf{r}).
    \label{eq:classic}
\end{align}
where $\mathbf{F}_s(\mathbf{x})$ represents spatial features encoded at a point position $\mathbf{x}$, and $\mathbf{r}$ is the reflected vector at point $\mathbf{x}$ for the viewing direction $\mathbf{v}$. This approximation works well if the incoming light is positionally invariant, but its accuracy degrades in the presence of local lighting effects such as close point lights, inter-reflections and self-shadows. In such cases, the spatial features must additionally encode some information about the local lighting and thus require a larger dimensionality than one would expect. A large and deep MLP is also necessary to decode this compressed information, which lowers the computational efficiency.

\begin{figure}[h]
    \centering
    \includegraphics[width=0.9\linewidth]{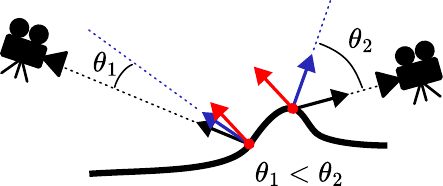}
    \caption{The reflectivity cannot be encoded in $\mathbf{F}_a$ alone because different \dif{incident vectors} can map to the same reflected vector (in red). In that case, similar angular features will be obtained for the two viewpoints due to the spatial and angular proximity of the lookups. The angle of incidence disambiguates the two situations.}
    \label{fig:view_angle}
\end{figure}

We propose to decouple the radiance model into spatial and angular features. We model a 3D scene with a combination of a high resolution sparse voxel grid that models the SDF field and a coarser grid that models the angular features as shown in fig. \ref{fig:overview}. For each point in space, the spatial features are obtained with a planar factorization in the coarse resolution tile with $\mathbf{F}_s = \mathbf{F}_x \otimes \mathbf{F}_y \otimes \mathbf{F}_z$ for dimensionality reduction \cite{Chen2022ECCV}. The angular features \dif{$\mathbf{F}_a$} are obtained by tri-linearly interpolating light field probes as explained in sec.~\ref{subsec:parametrization}. In addition, a dependency on the angle of incidence must be added to account for differences in reflectivity, as shown in fig. \ref{fig:view_angle}. 

The reflectivity, also called the Fresnel term, is commonly approximated as follows for rendering purposes \cite{Schlick1994AnIB}: 
\begin{align}
    R(\theta) = R_0 + (1 - R_0)(1- \cos{\theta}) ^ 5.
    \label{eq:shlick}
\end{align}
$R_0$ is the reflectivity at normal incidence and $\theta$ is the angle between the view direction and the half-way vector to the light source. In our case, $R_0$ is unknown and there is no explicit light source which means that we cannot use this formula directly. Instead, we approximate $\cos{\theta}$ by $\mathbf{n} \cdot \mathbf{v}$ and let the \dif{first layer of the} neural network learn a polynomial approximation of the Fresnel term. Our revised parametrization is given by eq. \ref{eq:newparametrization}:
\begin{multline}
    C = MLP(\mathbf{F}_s(\mathbf{x}), \mathbf{F}_a(\mathbf{x}, \mathbf{r}), \\ (1-\mathbf{n} \cdot \mathbf{v})^0, \dots, (1-\mathbf{n} \cdot \mathbf{v})^5).
    \label{eq:newparametrization}
\end{multline}
We denote the dimensionality of the features by $n_s$ and $n_a$ such that $\mathbf{F}_s \in \mathbb{R}^{n_s}$ and $\mathbf{F}_a \in \mathbb{R}^{n_a}$. Most materials reflect the light in a cone around the reflected vector which is why we condition $\mathbf{F}_a$ on $\mathbf{r}$, but some materials have retro-reflective properties that break this assumption. In that case, conditioning on the view direction would be more appropriate. Lastly, eq. \ref{eq:newparametrization} assumes isotropic materials since the reflectivity depends only on $\mathbf{n} \cdot \mathbf{v}$. Anisotropic materials could be handled by adding a dependency on the azimuthal angle.

\subsection{Light Field Probes on a Coarse Grid}
\label{subsec:parametrization}

\begin{figure}[h]
    \centering
    \includegraphics[width=1.0\linewidth]{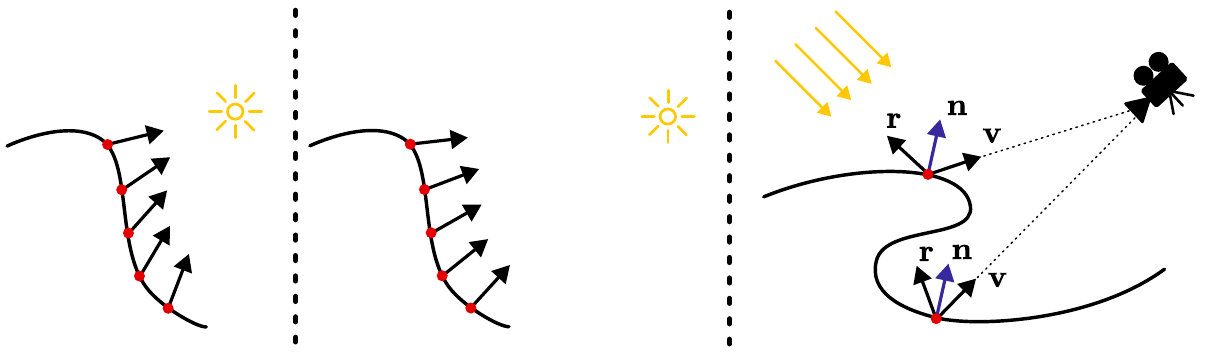}
    \caption{Left and center: The lighting direction changes at a lower rate on the surface for distant light sources. Right: Self shadows impose variations in lighting even with directional illumination.}
    \label{fig:lighting}
\end{figure}

In this work, we replace the generic angular inputs $\mathbf{v}$, $\mathbf{n}$, $\mathbf{r}$ with new angular inputs encoding the local illumination (the light field probes), with the goal to achieve a more efficient parametrization of the appearance in terms of parameter count, training and inference speed. Similarly to previous works \cite{yu2022plenoxels, kerbl3Dgaussians}, we use a spherical harmonics decomposition to model this dependency, but with several important differences: first, we encode abstract features on the sphere rather than the outgoing radiance which lets a neural network handle the non-linearities. Second, our directional embedding has a low spatial resolution which takes advantage of the smooth spatial variations of the lighting. Third, we sample this embedding with the reflected vector instead of the view vector to link the local surface orientation with the shading function, which has been showed to be beneficial for improved surface quality~ \cite{verbin2022refnerf}. The angular features encoded into a single neural probe are given by eq. \ref{eq:spherical_harmonics}:
\begin{align}
    \mathbf{F}_a = \mathcal{SH}(\mathbf{r}) = \sum_{j=1}^{l^2} \mathbf{b}_j Y_j\left(\mathbf{r}\right).
    \label{eq:spherical_harmonics}
\end{align}
The polynomials of the SH basis are denoted with $Y_j$. There are $l^2$ coefficients\footnote{We use a one-indexed notation for the SH order $l$ instead of the traditional zero-indexing so that we have $l^2$ coefficients.} $\mathbf{b}_j \in \mathbb{R}^{n_a}$. Increasing values of $l$ unlock signals of increasing angular frequency which gives an explicit control on the representable amount of specularities. We observe two guiding principles: (1) The shinier the material, the higher the frequency of $\mathbf{F}_a$ should be in the angular domain. (2) The closer the light source, the higher the frequency $\mathbf{F}_a$ should be in the spatial domain. See fig. \ref{fig:lighting} for an illustration.

For the case of common scenes such as humans or everyday objects, assuming materials with high to medium roughness so that $\mathbf{F}_a$ does not need to be of very high angular frequency is generally sufficient.  We also assume that the lights are far enough, so that probes at a low spatial frequency can explain the parallax.
The probes are stored at 1/16th the resolution of the main voxel grid, as shown in fig. \ref{fig:overview}. The voxels tri-linearly interpolate the 8 nearest probes according to eq. \ref{eq:sh_eval1} and \ref{eq:sh_eval2}, with $w_i$ the interpolation weight of the i-th corner.
\begin{align}
    \mathbf{F}_a &= \sum_{i=1}^8 \mathcal{SH}_i(\mathbf{r}) w_i = \sum_{i=1}^8 \sum_{j=1}^{l^2} \mathbf{b}_{ij} Y_j(\mathbf{r}) w_i \label{eq:sh_eval1}\\
    &=\sum_{j=1}^{l^2} \hat{\mathbf{b}}_{j} Y_j(\mathbf{r}) \quad \text{with } \hat{\mathbf{b}}_{j} = \sum_{i=1}^8 \mathbf{b}_{ij} w_i \label{eq:sh_eval2}
\end{align}

\subsection{SDF Estimation Approach}
\label{subsec:regs_and_losses}

We here discuss how to embed the new proposed color decoder in an implicit surface reconstruction pipeline. During the forward pass, for a pixel $(u,v)$ in a given viewpoint, the color is computed by a transmittance $T_i$-weighted average of the colors $C_i$ of $N$ points sampled from the current scene volume estimate along the pixel aligned ray, where $T_i$ is the product of obstructing opacities $\alpha_j$. We use the NeuS equation to relate volumetric opacities to the SDF $s_i$ at each sample point~\cite{wang2021neus}:
\begin{align}
    &c(u,v) = \sum_{i=1}^N T_i\alpha_iC_i, \qquad T_i = \prod_{j < i} (1-\alpha_j), \\
    &\alpha_i = \mathrm{max}\left(\frac{\Phi_{\tau}(s_{i})-\Phi_{\tau}(s_{i+1})}{\Phi_{\tau}(s_{i})}, 0\right), \\
    &\Phi_{\tau} (s_{i}) = \frac{1}{1+\exp^{-\tau s_{i}}}.
    \label{eq:neus}
\end{align}
where $\tau$ is a scale factor that increases during optimization. The SDF field can then be optimized by backpropagating updates from the observed color loss through the rendering equation, and from the set of regularization losses. 

Our losses and regularizations are detailed in equations~ \ref{eq:smooth_sdf} to \ref{eq:photo_loss}. The raw SDF $\hat{s}$ is convolved by a $5^3$ gaussian kernel $G$ into a smoother SDF $s$ to stabilize training \cite{wu2022voxurf}. Note that $s$ is used for all subsequent operations such as normal computation and rendering. Eq. \ref{eq:smooth_sdf} maintains $s$ and $\hat{s}$ close to each other to promote smoothness. The Eikonal regularization \ref{eq:eikonal_reg} ensures that $s$ is a signed distance. Equations \ref{eq:smmoth_normals}, \ref{eq:smooth_features}, \ref{eq:smooth_probes} promote spatial smoothness for the normals, for the factorized spatial features and for the probe coefficients respectively.
In eq. \ref{eq:smooth_probes}, $\mathcal{V}_i$ is the set of neighboring probes adjacent to the i-th one and $N$ is the total number of probes. The regularizations are applied on each voxel, factorized feature, probe or image pixel when appropriate.
\begin{align}
    \mathcal{L}_{\text{sdf}} &= \sum_\text{voxel} |s - \hat{s}|^2, \quad s = G(\hat{s}) \label{eq:smooth_sdf} \\
    \mathcal{L}_{\text{Eik}} &= \sum_\text{voxel} (||\nabla s|| - 1)^2  \label{eq:eikonal_reg}\\
    \mathcal{L}_{\text{normal}} &= \sum_\text{voxel} ||\nabla \mathbf{n}||^2, \quad \mathbf{n} = \nabla s / || \nabla s || \label{eq:smmoth_normals} \\
    \mathcal{L}_{\text{features}} &= \sum_\text{texel} ||\nabla \mathbf{F}_x||^2 + ||\nabla \mathbf{F}_y||^2 + ||\nabla \mathbf{F}_z||^2 \label{eq:smooth_features} \\
    \mathcal{L}_\text{probes} &= \sum_{i=1}^N \sum_{j=1}^{l^2} \sum_{k \in \mathcal{V}_i} ||\mathbf{b}_{ij} - \mathbf{b}_{kj} ||^2 \label{eq:smooth_probes} \\
    \mathcal{L}_\text{photo} &= \sum_{u,v} || c(u,v) - c_{\text{gt}}(u,v) ||^2 \label{eq:photo_loss}
\end{align}
In eq. \ref{eq:photo_loss}, $c$ is the rendered image and $c_\text{gt}$ is the ground truth image. We apply a weight of $(\text{max}(c,c_\text{gt}) + \epsilon)^{-1}$  on the photometric gradient of each pixel in eq. \ref{eq:photo_loss} to penalize relative differences rather than absolute differences \cite{Mildenhall2021NeRFIT}. \dif{A similar strategy is used on $\mathcal{L}_\text{sdf}$ and $\mathcal{L}_\text{features}$.} We apply an  empirical factor of $(1 + |s| * 5)^{-1}$ on the gradients of the per-voxel regularizations to lower their importance \dif{when the distance to the zero-crossing increases}. The idea is to keep the regularizations and the data term balanced \dif{for regions near and far from the surface alike}. 

\subsection{Training}
\label{subsec:training}

Our complete loss is a weighted sum of eq. \ref{eq:smooth_sdf} to \ref{eq:photo_loss} that we minimize with the Adam optimizer \cite{Kingma2014AdamAM}, \dif{please refer to the supplementary document for more details}. In summary, the trained parameters are the raw SDF values $\hat{s}$, the factorized features $\mathbf{F}_x, \mathbf{F}_y, \mathbf{F}_z$, the probes coefficients $\mathbf{b}$ and the MLP weights.
We follow a coarse-to-fine learning strategy, both in the spatial and angular domains. We start with coarse voxels and $l=2$, under the supervision of low resolution images. We iteratively subdivide the voxels and switch to higher resolution images, and introduce higher angular frequencies by increasing $l$. Starting the optimization with too many degrees of freedom for the appearance parametrization can lead to local minima, as noted by \cite{kerbl3Dgaussians}.

\section{Implementation Details}

We use a 2 hidden layer MLP with 32 neurons and ReLU activations with a sigmoid for the last layer. The probes at the shared corners of adjacent tiles are duplicated in memory (but keep shared values) to benefit from hardware-accelerated interpolation. We support up to 16 spherical harmonic coefficients ($l=4$) per probe. The computation of $\mathbf{F}_a$, $\mathbf{F}_s$ and the MLP call are \dif{fused into} a single CUDA kernel for efficiency, following \cite{mueller2022instant} and \cite{millimetrichumans}. Each $16^3$ tile contains $3\times16\times16\times n_s$ spatial coefficients and $8\times l^2\times n_a$ spherical harmonic features. Assuming $l=4$ and $n_s=n_a$, the probes only require $(8\times16) / (3\times16\times16) = 1/6\text{th}$ of the memory taken by the spatial features. The sparse voxel grid is based on the code of \cite{millimetrichumans}, with targeted modifications to implement our probes.

\section{Experiments}
\label{subsec:experiments}

\textbf{Protocol} We evaluate our method on 4 datasets : MVMannequins \cite{millimetrichumans}, ActorsHQ \cite{isik2023humanrf}, DTU \cite{jensen2014large} and BlendedMVS \cite{yao2020blendedmvs}. 
\dif{Due to the various evaluation settings employed in the community, we chose to re-train and re-evaluate all the baselines using the same experimental protocol for fairness. Following \cite{Huang2DGS2024}, all available images were given as input and used for evaluation. Timings are measured on a workstation equipped with an RTX A6000 GPU (Ampere series). PSNR values are computed inside the silhouette region of objects of interest and chamfer error is computed using official scripts. We use the following grid resolutions: DTU ($1000^3$), BMVS ($666^3$), MVMannequins ($2500\times1250\times2500$, 2mm/voxel), ActorsHQ ($3000\times4000\times3000$, 0.6mm/voxel). Some datasets may contain inconsistent shadows or large exposure changes (DTU and BMVS) that are hard to model by our representation so we optionally train per-camera bias vectors in the MLP.}

\begin{table*}[htb]
	\begin{center}
	    \small
		\begin{tabular}{|c|c|c|c|c|c|c|c|c|c|c|}
			\hline
			Metrics & (4,4,4)\xmark & (8,8,4)\xmark & MMH \cite{millimetrichumans} & Voxurf \cite{wu2022voxurf} & Neus2 \cite{neus2} & Colmap \cite{schoenberger2016sfm} & 2DGS \cite{Huang2DGS2024} \\
			\hline
			Chamfer (mm) & \underline{1.04} & \textbf{1.03} & 1.14 & 1.59 & 2.13 & 3.52 & 3.35 \\
			\hline
			PSNR & \underline{36.81} & \textbf{36.90} & 36.33 & 35.51 & 34.22 & - & 34.89 \\
			\hline
			Training & 1min & 1min & 2-3min & 15min & 5min & \textgreater 1h & \textgreater 1h \\
			\hline
		\end{tabular}
	\caption{\dif{Evaluation on MVMannequins with two configurations for $(n_s, n_a, l)$. Per-camera bias vectors are not used (\xmark-symbol).}}
	\label{tab:mvmannequins}
	\end{center}
\end{table*}

\textbf{MVMannequins} is a dataset of 14 dressed mannequins for multi-view reconstruction benchmarking with 68 cameras at $2048^2$ resolution. We use the official evaluation code and report the metrics in table \ref{tab:mvmannequins}. We outperform all the baselines while taking only about a minute of training time, almost 3x faster than \cite{millimetrichumans}, with 5GB of VRAM, a rendering speed of 300Hz to 400Hz and a model size of 30MB.

\begin{table}[htb]
	\begin{center}
	    \small
		\begin{tabular}{|c|c|c|c||c|c|}
		    \hline
			\multicolumn{1}{|c|}{} & \multicolumn{3}{c||}{$(n_s, n_a, l)=(4,4,4)$\xmark} & \multicolumn{1}{c|}{Voxurf} & \multicolumn{1}{c|}{NeuS2} \\
			\hline
			 Resolution & r/1 & r/2 & r/4 & r/2 & r/2\\
			\hline
			PSNR & \textbf{37.48} & \underline{36.62} & 34.74 & 36.56 & 34.53 \\
			\hline
			\hline
			Training & 250s & 110s & 53s & 1h & 250s \\
			\hline
			Model size & 232MB & 57MB & 15MB & 500MB & 25MB \\
			\hline
		\end{tabular}
    	\caption{\dif{Evaluation for ActorsHQ. Per-camera bias vectors are not used (\xmark-symbol). Full, half and quarter-resolution input images are denoted as $r/1$, $r/2$, $r/4$ respectively. Renderings are up-scaled before comparison with the full resolution images ($4088\times2990$).}}
    	\label{tab:actorshq}
	\end{center}
	\begin{center}
	\small
	\begin{tabular}{|c|c|c|c|c|c|}
		\hline
		Metrics & (4,4,4)\cmark & (8,8,4)\cmark & Voxurf & Neus2 & 2DGS \\
		\hline
		Chamfer & \textbf{0.68} & \underline{0.71} & 0.73 & 0.80 & 0.76 \\
		\hline
		PSNR & 37.03 & \textbf{37.74} & \underline{37.08} & 36 & 36.03 \\
		\hline
		\hline
		Training & 150s & 160s &  15min & 5min & 12min \\
		\hline
		Model size & 56MB & 88MB & 500MB & 25MB & 49MB \\ 
		\hline
	\end{tabular}
	\caption{\dif{DTU evaluation with two configurations for $(n_s, n_a, l)$. Per-camera bias vectors are used (\cmark-symbol).}}
    \label{tab:DTU}	
	\end{center}
	\begin{center}
	\small
		\begin{tabular}{|c|c|c|c|c|}
			\hline
			Metrics & (4,4,4)\cmark & (8,8,4)\cmark & Voxurf & Neus2 \\
			\hline
			Chamfer & \underline{2.37} & \textbf{2.22} & 2.64 & 2.93 \\
			\hline
			PSNR & \underline{35.19} & \textbf{35.89} & 35.11 & 33.62 \\
			\hline
			\hline
			Training & 110s & 120s & 15min & 5min \\
			\hline
		\end{tabular}
	\caption{\dif{BMVS evaluation with two configurations for $(n_s, n_a, l)$. Per-camera bias vectors are used (\cmark-symbol).}}
	\label{tab:BMVS}
	\end{center}
\end{table}

\begin{figure}[htb]
    \centering
    \includegraphics[width=1.0\linewidth]{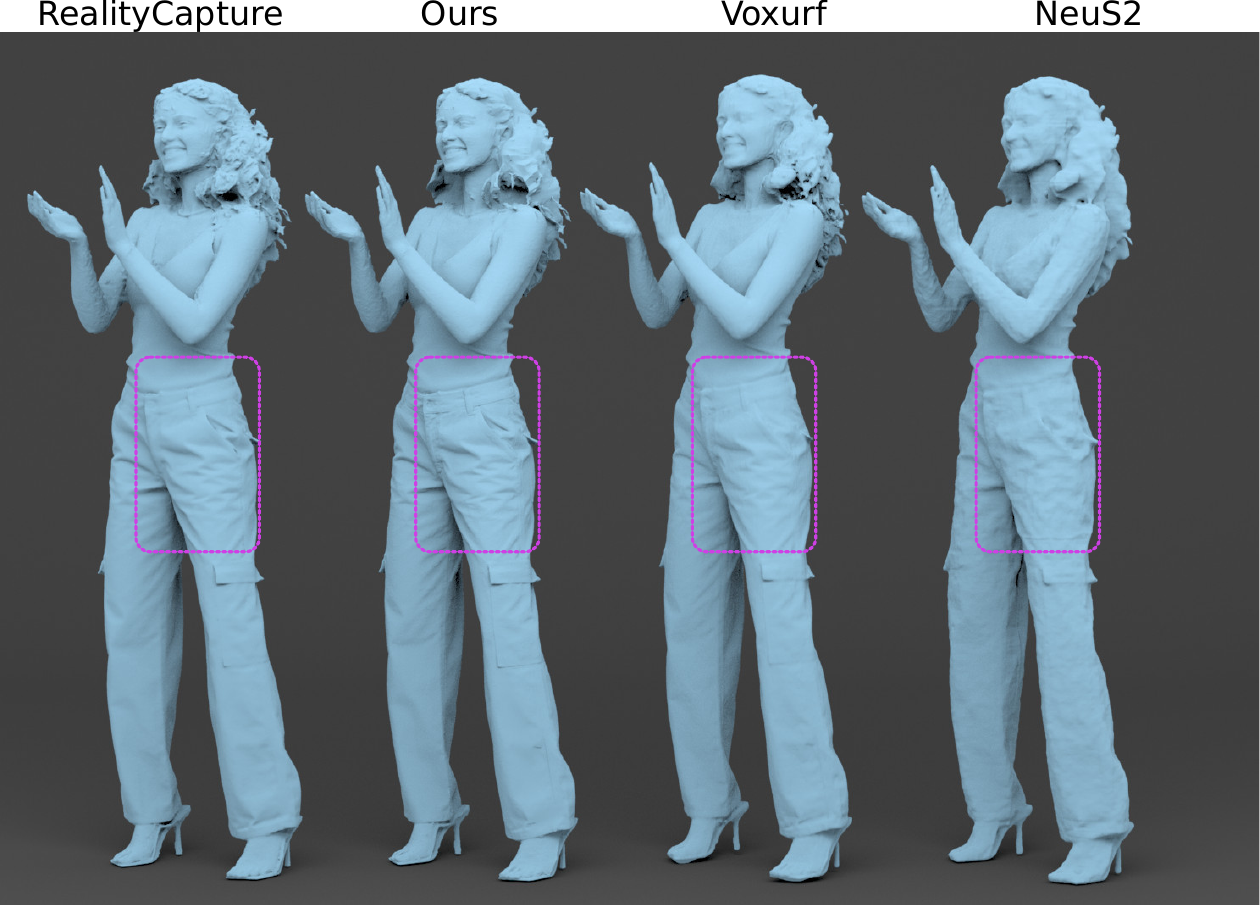}
    \includegraphics[width=1.0\linewidth]{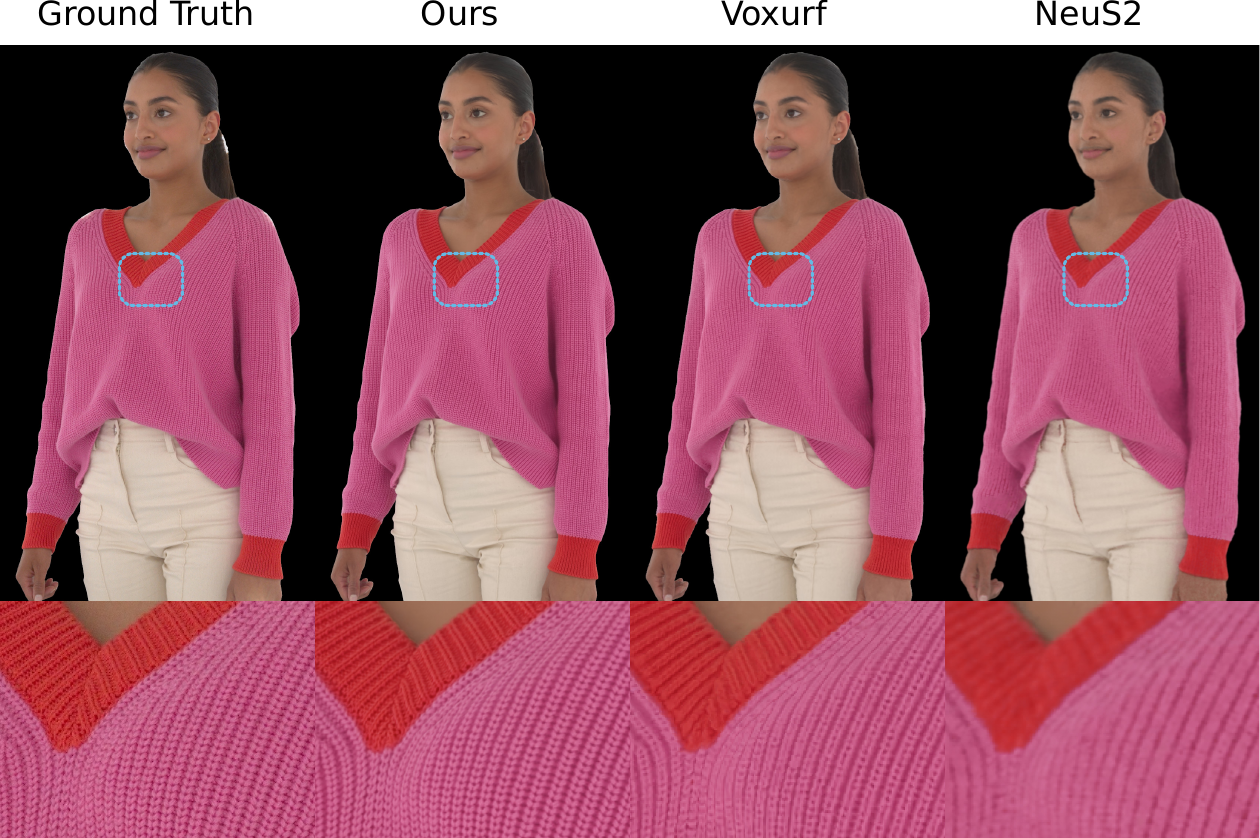}
    \caption{High-resolution reconstructions for ActorsHQ.}
    \label{fig:ActorsHQVisu}
\end{figure}

    \textbf{ActorsHQ} is a multiview dataset of humans in motion with 160 cameras at a high resolution of $4088 \times 2990$. We evaluate on the 1000th temporal frame of 13 sequences. We compare against Voxurf \cite{wu2022voxurf}  and NeuS2 \cite{neus2}, which we trained with half resolution images since we ran into issues with full resolution images. We report the PSNR in table \ref{tab:actorshq} \dif{and show qualitative results in fig. \ref{fig:ActorsHQVisu}}. Voxurf takes about one hour to fully converge (with $3\times$ the training iterations of the DTU configuration). NeuS2 converges fast but with a lower PSNR. In comparison, our method only takes $\sim$4 minutes to converge on the full resolution images, and we also outperform both when training at the same resolution. \dif{This benchmark highlights the capability of our method to gracefully scale to high-resolution scenes.}

\begin{figure}[htb]
\includegraphics[width=\linewidth]{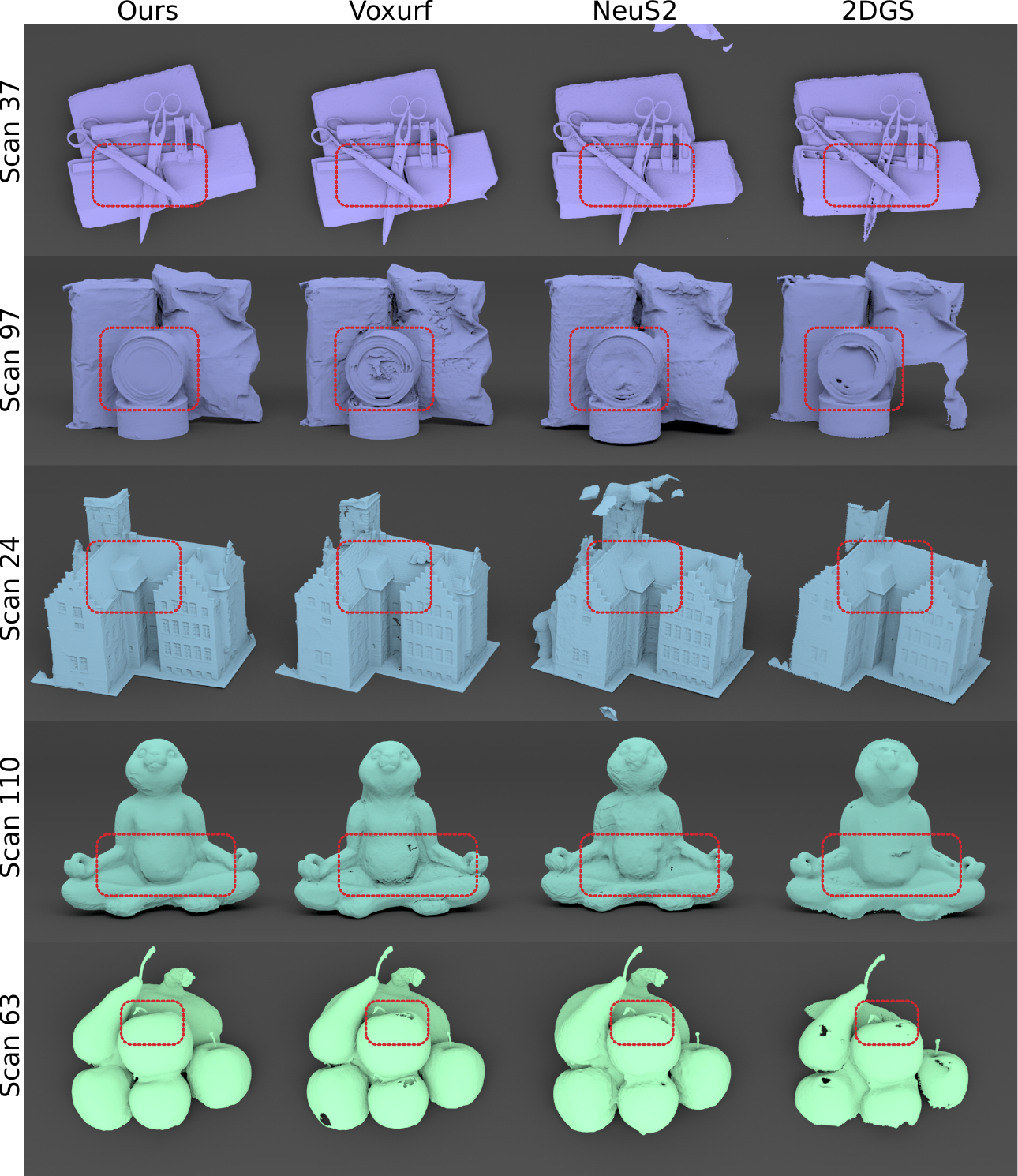}
\caption{Reconstructions on challenging examples of DTU.}
\label{fig:DTUVisu}
\end{figure}

\begin{figure}[htb]
\includegraphics[width=\linewidth]{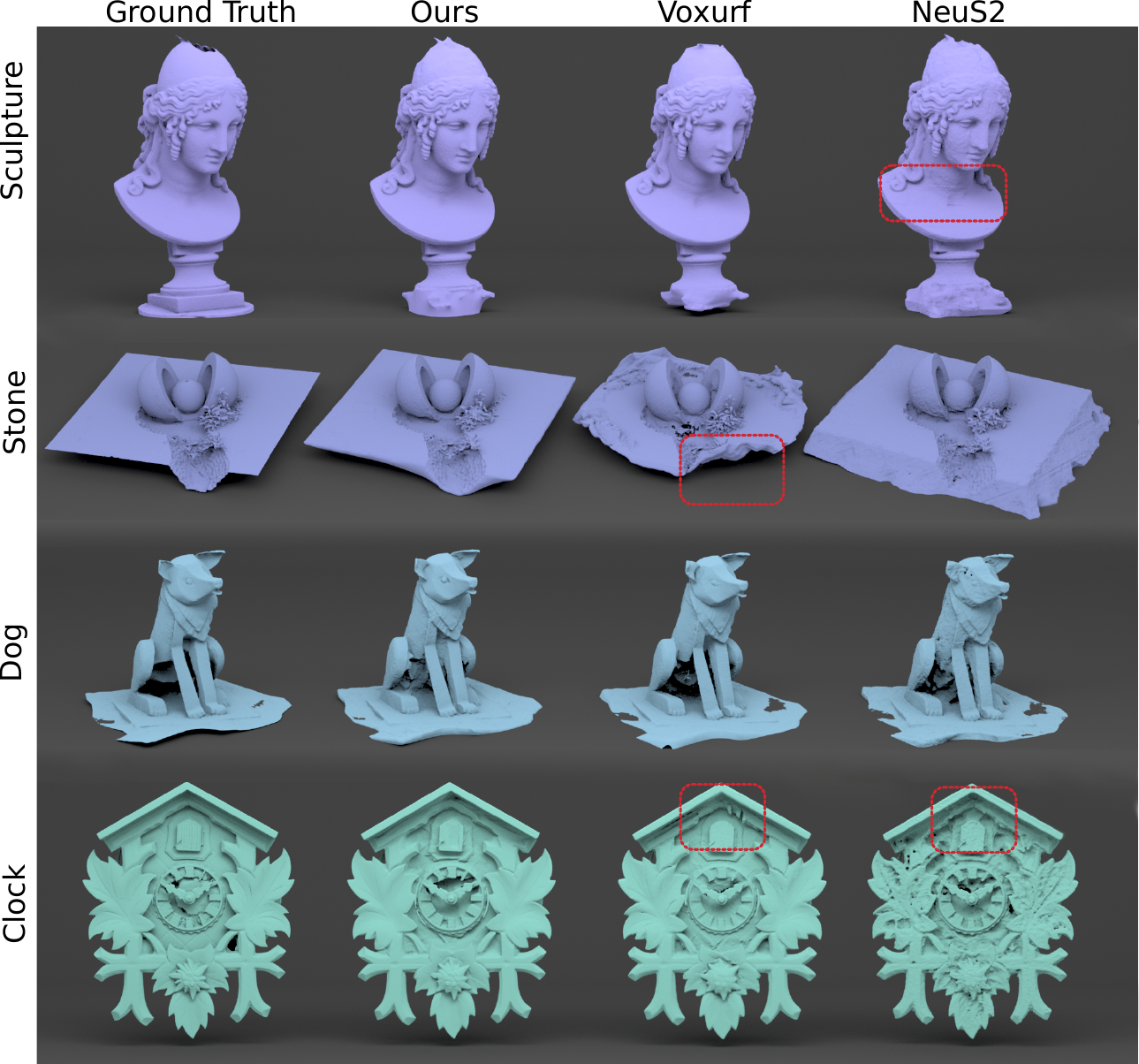}
\caption{Reconstructions examples on BlendedMVS.}
\label{fig:bmvs_geom_comp}
\centering
\end{figure}

\textbf{DTU} provides $49$ to $64$ images of various objects at $1600\times1200$ resolution, along with evaluation point clouds. We use the foreground masks provided by IDR \cite{yariv2020multiview} for training and testing. 2DGS \cite{Huang2DGS2024} does not support masks for training so it uses complete images instead, cropped and at half resolution according to its evaluation protocol. We compare against Voxurf \cite{wu2022voxurf}, Neus2 \cite{neus2} and 2DGS \cite{Huang2DGS2024} on DTU in table \ref{tab:DTU}, using the official python script for geometric evaluation. We outperform all three in terms of chamfer distance and PSNR, twice as fast as NeuS2 \dif{on the same hardware}. Rendering speed ranges from 200Hz to 300Hz.

\textbf{BlendedMVS.} Finally, we test our method on 8 scenes of the BlendedMVS dataset \cite{yao2020blendedmvs} and report the metrics in table \ref{tab:BMVS}. The BlendedMVS is a semi-synthetic dataset were objects have been first reconstructed then reprojected in the images in order to produce reliable ground truth geometry and masks. There is no official script for geometric evaluation so we measure two-ways point to mesh distances in the reference frame used for training in \cite{wang2021neus}, scaled by 1000x.

\begin{figure}[htb]
    \centering
    \includegraphics[width=1.0\linewidth]{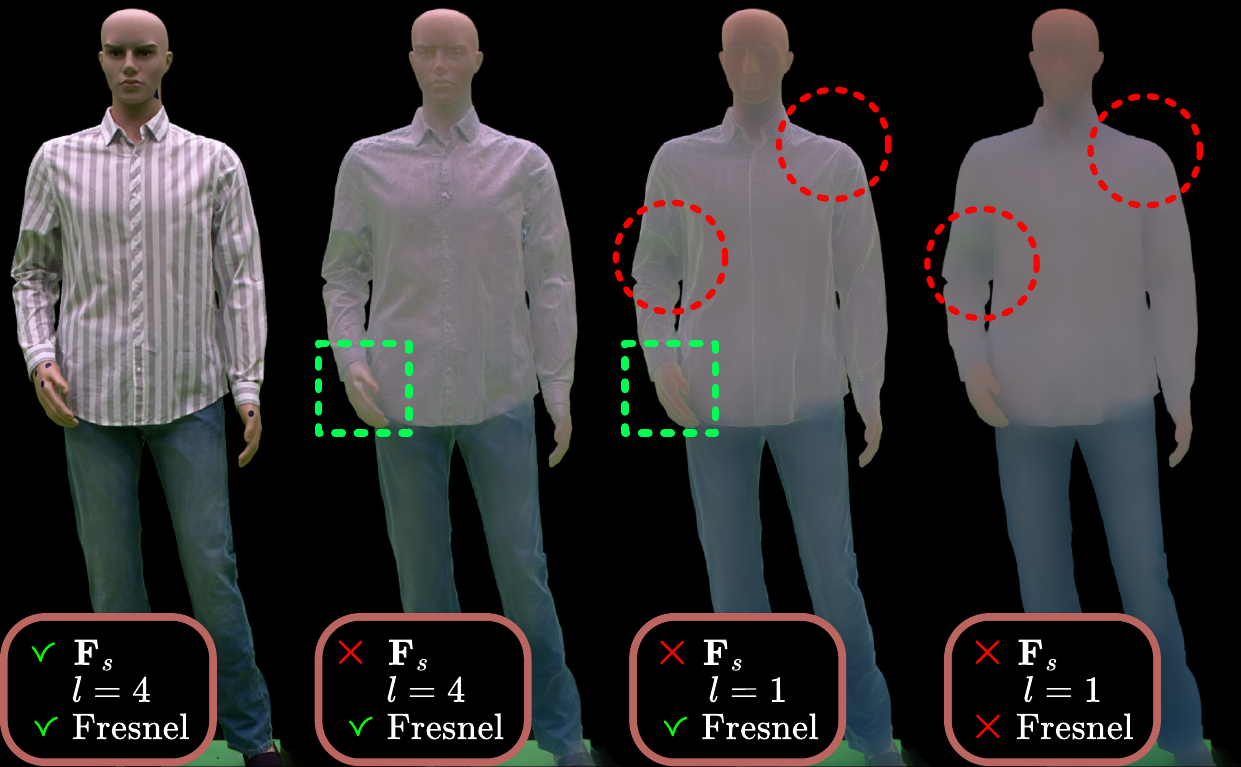}
    \caption{\dif{Our model can be intuitively explained. We disable some components of a pre-trained example from the MVMannequins dataset \cite{millimetrichumans} (left). From left to right, we remove the spatial features (set $\mathbf{F}_s=0$), only keep the constant coefficient of the SH basis ($l=1$), disable the incidence angle embedding (set $\mathbf{n} \cdot \mathbf{v}=1$ in eq. \ref{eq:newparametrization}). The learned Fresnel term encodes grazing reflections (red circles), the angular features provide the shading (green squares) and the spatial features contain the high frequency details.}}
    \label{fig:MLPCompenentsAblation}
\end{figure}

\begin{figure}[htb]
    \centering
    \includegraphics[width=1.0\linewidth]{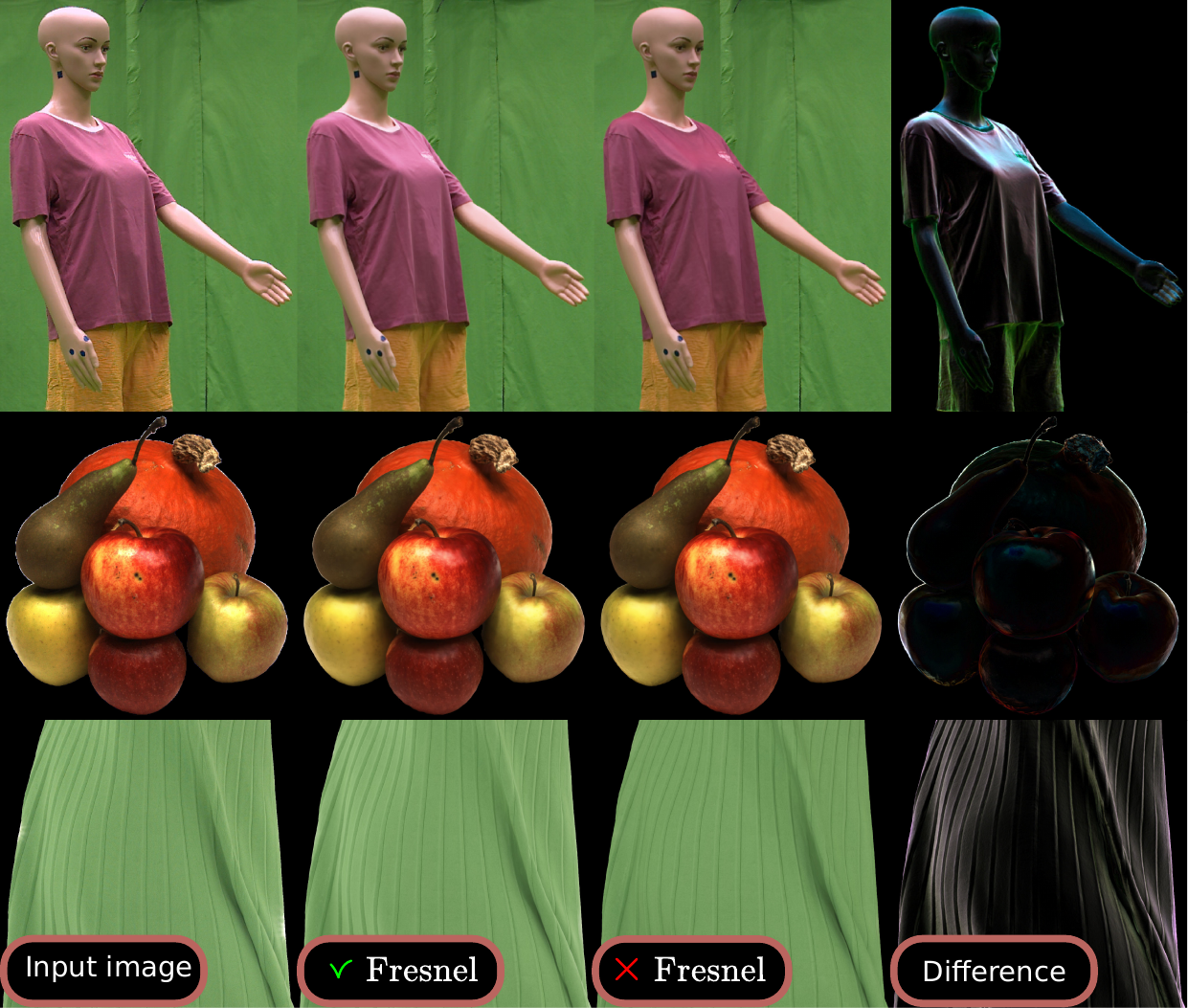}    \caption{\dif{Effect of the learned Fresnel term on three examples from MVMannequins, DTU and ActorsHQ. Left to right: input image, volume rendering (A), volume rendering w/o the Fresnel term (B) (we set $\mathbf{n} \cdot \mathbf{v}=1$ in eq. \ref{eq:newparametrization}), difference image ($4\times|A-B|$).}}
    \label{fig:Fresnel}
\end{figure}

\textbf{Ablations.} \dif{We ablate the probes smoothing and the polynomial approximation of the Fresnel term in tab. \ref{tab:ablation} on the MVMannequins dataset by imposing $\mathbf{n} \cdot \mathbf{v}=1$ in eq. \ref{eq:newparametrization} during training, with noticeable incidence on Chamfer error. A visual ablation of the components of our parametrization is presented in fig. \ref{fig:MLPCompenentsAblation} on the \textit{kino/jea} mannequin, clearly showing the role of each parameter group. Figure \ref{fig:Fresnel} contains more examples. Various choices of configurations are compared in table \ref{tab:bmvs_ablations} on the BlendedMVS dataset.}
\begin{table}[htb]
    \centering
    \scriptsize
    \begin{tabular}{|c|c|c|c|}
        \hline
         & (4,4,4)\xmark & w/o probes smoothing & w/o Fresnel \\
        \hline
        Chamfer (mm) & \textbf{1.04} & \underline{1.09} & 1.18 \\
        \hline
        PSNR & \underline{36.81} & \textbf{36.91} & 36.79 \\
        \hline
    \end{tabular}
    \caption{\footnotesize Ablation on MVMannequins}
    \label{tab:ablation}
	\begin{center}
		\small	
		\resizebox{\columnwidth}{!}{%
		\begin{tabular}{|c|c|c|c|c|c|c|c|}
			\hline
			Metrics & (4,4,4)\xmark & (4,4,4)\cmark & (8,8,4)\cmark & (12,12,4)\cmark & (4,4,1)\cmark & (4,4,2)\cmark & (4,4,3)\cmark \\
			\hline
			Chamfer & 2.47 & 2.37 & \textbf{2.22} & \underline{2.31} & 2.58 & 2.34 & 2.37 \\
			\hline
			PSNR & 34.76 & 35.19 & \underline{35.89} & \textbf{36.17} & 34.44 & 34.86 & 35.05 \\
			\hline
		\end{tabular}}
	\caption{We ablate the number of spatial features $n_s$, angular features $n_a$ and spherical harmonics order $l$ on BlendedMVS. Each triplet denotes a configuration $(n_s, n_a, l)$. The \cmark-symbol indicates when per-camera bias vectors are trained.}
    \label{tab:bmvs_ablations}
	\end{center}
\end{table}


\section{Limitations}
\label{sec:limitations}

We identify several limitations to our approach. First, the fact that the lighting is encoded locally can degrade the extrapolation capabilities compared to modeling the environment with a global representation. This becomes apparent on the ActorsHQ dataset where there is a good overlap of the cameras field of views along the longitude, but very little overlap along the latitude on the legs. This creates shadows artifacts when looking from above or below rather that at a level angle. Second, although our decomposition is physically-inspired, we do not explicitly simulate light transport which can result in local minima due to shape-radiance ambiguities. \dif{Relighting and/or material editing is not supported but fine-tuning the probes to new lighting seems possible and worthy of future investigations.}

\section{Conclusion and Future Work}
\label{sec:conclusion}

We have presented a novel approach to model the view dependent appearance, enabling both fast reconstruction and real-time rendering of humans and objects. Notably, our model does not compromise on geometric and photometric quality, even surpassing the \dif{previous approaches} in many instances. It can easily scale to high resolution inputs while maintaining high performance. We have reached a parsimonious parametrization of the surface properties thanks to a simple formulation. Disentangling the spatial and angular components with high efficiency also offers promising future avenues of research toward temporal reconstruction. Other future research can build on this work, \eg handling high frequency reflections at a lower cost than current dedicated approaches and \dif{unifying} neural rendering and physically-based rendering. Other embeddings of the angular dependency representations can be investigated, such as cubemaps or mixtures of spherical gaussians. 

\section{Acknowledgments}

This work was supported by French government funding managed by the National Research Agency under the
Investments for the Future program (PIA) grant ANR-
21-ESRE-0030 (CONTINUUM). Research conducted at
Kyushu University was supported by JSPS/KAKENHI
JP23H03439 and AMED JP24wm0625404. We thank Laurence Boissieux for her valuable help generating 3D renderings, and Edmond Boyer for insightful discussions and comments.

{
    \small
    \bibliographystyle{ieeenat_fullname}
    \bibliography{main}
}

\appendix
\clearpage
\setcounter{page}{1}
\maketitlesupplementary

\section{Scheduling}

\begin{table*}[htb]
    \centering
    \resizebox{\textwidth}{!}{%
    \begin{tabular}{|c|c|c|c|c|c|c|c|c|c|c|c|c|}
        \hline
        \makecell{Level of\\detail} & \makecell{Learning rate\\(Voxels)} & \makecell{Learning rate\\(MLP)} & $\lambda_\text{Eik}$ & $\lambda_\text{sdf}$ & $\lambda_\text{features}$ & $\lambda_\text{normal}$ & $\lambda_\text{probes}$ & \makecell{Images per\\batch} & Iterations\\
        \hline
        LOD 4 & [0.025, 0.01] & 0.01 & [1.0, 0.1, 0.1] & [2.0, 0.2, 0.2] & [0.5, 0.05, 0.05] & [0.5, 0.05, 0.05] & [2.0, 0.2, 0.2] & 8 & 3000\\
        \hline
        LOD 3 & & & [0.2, 0.1] & [0.4, 0.2] & [0.1, 0.05] & [0.1, 0.05] & [0.4, 0.2] & &\\
        \hline
        LOD 2 & & & & & & & & &\\
        \hline
        LOD 1 & & & & & & & & & 1000\\
        \hline
        LOD 0 & [0.01, 0.001] & [0.01, 0.001] & & & & & & 4 & 500 \\
        \hline
    \end{tabular}}
    \caption{Schedule for DTU}
    \label{tab:schedule_dtu}
    \centering
    \resizebox{\textwidth}{!}{%
    \begin{tabular}{|c|c|c|c|c|c|c|c|c|c|c|c|c|}
        \hline
        \makecell{Level of\\detail} & \makecell{Learning rate\\(Voxels)} & \makecell{Learning rate\\(MLP)} & $\lambda_\text{Eik}$ & $\lambda_\text{sdf}$ & $\lambda_\text{features}$ & $\lambda_\text{normal}$ & $\lambda_\text{probes}$ & \makecell{Images per\\batch} & Iterations\\
        \hline
        LOD 4 & [0.1, 0.01] & 0.01 & [0.25, 0.025, 0.025] & [0.5, 0.05, 0.05] & [0.25, 0.025, 0.025] & [0.025, 0.0025, 0.0025] & [2.0, 0.2, 0.2] & 8 & 4000\\
        \hline
        LOD 3 & & & [0.05, 0.025] & [0.1, 0.05] & [0.05, 0.025] & [0.005, 0.0025] & [0.4, 0.2] & &\\
        \hline
        LOD 2 & [0.025, 0.01] & & & & & & & & 3000\\
        \hline
        LOD 1 & & & & & & & & & 1500\\
        \hline
        LOD 0 & [0.01, 0.001] & [0.01, 0.001] & & & & & & 4 & 1000 \\
        \hline
    \end{tabular}}
    \caption{Schedule for BlendedMVS}
    \label{tab:schedule_bmvs}
    \centering
    \resizebox{\textwidth}{!}{%
    \begin{tabular}{|c|c|c|c|c|c|c|c|c|c|c|c|c|}
        \hline
        \makecell{Level of\\detail} & \makecell{Learning rate\\(Voxels)} & \makecell{Learning rate\\(MLP)} & $\lambda_\text{Eik}$ & $\lambda_\text{sdf}$ & $\lambda_\text{features}$ & $\lambda_\text{normal}$ & $\lambda_\text{probes}$ & \makecell{Images per\\batch} & Iterations\\
        \hline
        LOD 4 & [0.025, 0.01] & 0.01 & [0.05, 0.025] & [0.2, 0.1] & [0.1, 0.05] & 0 & [0.4, 0.2] & 4 & 1500\\
        \hline
        LOD 3 & & & & & & & & & \\
        \hline
        LOD 2 & & & & & & & & & \\
        \hline
        LOD 1 & & & & & & & & & \\
        \hline
        LOD 0 & [0.01, 0.001] & [0.01, 0.001] & 0.0125 & 0.05 & 0.025 & 0.0063 & 0.1 & & 1000 \\
        \hline
    \end{tabular}}
    \caption{Schedule for MVMannequins}
    \label{tab:schedule_mvmannequins}
    \centering
    \resizebox{\textwidth}{!}{%
    \begin{tabular}{|c|c|c|c|c|c|c|c|c|c|}
        \hline
        \makecell{Level of\\detail} & \makecell{Learning rate\\(Voxels)} & \makecell{Learning rate\\(MLP)} & $\lambda_\text{Eik}$ & $\lambda_\text{sdf}$ & $\lambda_\text{features}$ & $\lambda_\text{normal}$ & $\lambda_\text{probes}$ & \makecell{Images per\\batch} & Iterations\\
        \hline
        LOD 6 & [0.05, 0.025] & 0.01 & [0.05, 0.025] & [0.2, 0.1] & [0.025, 0.0125] & 0 & [0.4,0.2] & 8 & 500\\
        \hline
        LOD 5 & & & & & & & & & 1000\\
        \hline
        LOD 4 & & & & & & & & & 1500\\
        \hline
        LOD 3 & & & & & & & & & \\
        \hline
        LOD 2 & & & & & & [0.025, 0.0125] & & & 1000\\
        \hline
        LOD 1 & [0.01, 0.0025] & [0.01, 0.001] & 0.0125 & 0.05 & 0.0063 & 0.0125 & 0.1 & 4 & \\
        \hline
        LOD 0 & [0.01, 0.001] & 0.001 & & & & 0.025 & & & 500 \\
        \hline
    \end{tabular}}
    \caption{Schedule for ActorsHQ}
    \label{tab:schedule_actorshq}
\end{table*}

The scheduling of the hyper-parameters is summarized in table \ref{tab:schedule_dtu}. We used $\beta_1=0.9$, $\beta_2 = 0.995$ and $\lambda_\text{photo}=40$ in all experiments, where $\beta_1$ and $\beta_2$ are ADAM's first and second moments. A blank cell indicates that the value is identical to that of the row above. Values between brackets are linearly interpolated based on the current training iteration of the given level of detail. Our general strategy is to start with a high learning rate and high regularizations that are progressively halved during training. We lower the learning rate in the final level of detail to stabilize convergence. The learning rate is also linearly increased starting from zero during the first 50 iterations of each level of detail as a warm-up. We use stronger regularizations at the coarsest level of detail on DTU and BMVS since the initialization from a sphere is poorer compared to the visual hull initialization of the other two datasets. The gradients are accumulated over 4 to 8 \textit{complete} images before stepping the optimizer. The photometric loss is divided by the number of images in the batch as a normalization. We adjust the number of levels of detail based on the resolution of the input images. The grid resolution is manually chosen so that a voxel roughly projects to an area equivalent to that of a pixel. The MLP weights and probe features are trained with a lower learning rate compared to the spatial features and voxel sdf values, also for stability.

\section{Visual ablation}

\begin{figure}[htb]
\includegraphics[width=\linewidth]{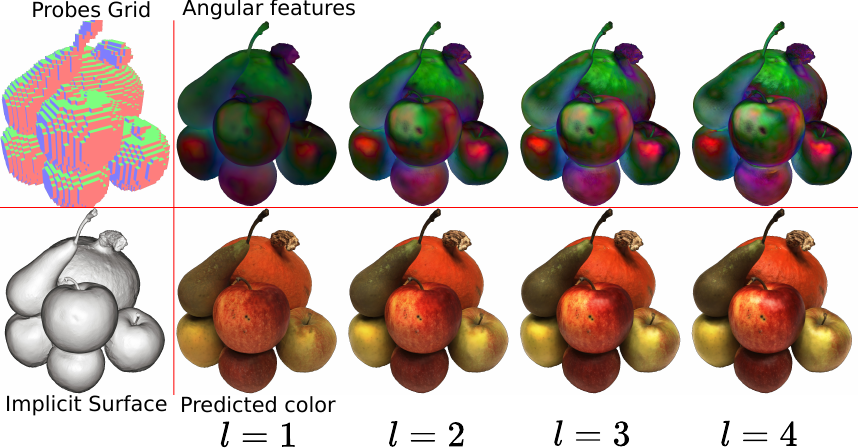}
\caption{We train a scene with $l=4$, that we then visualize with $l$ varying from 1 to 4, left to right. The corresponding angular features are on the top row, and the predicted color is on the bottom row. We observe that the specularities can be removed simply by disabling the high order spherical harmonics coefficients.}
\label{fig:ablations_l}
\end{figure}

A visual ablation of the spherical harmonic order $l$ is presented in fig \ref{fig:ablations_l}.

\section{Detailed tables}

Tables \ref{tab:mvmannequins_chamfer} to \ref{tab:bmvs_ablations_psnr} contain detailed metrics from all our experiments. The dimensionality of the spatial features $\mathbf{F}_s$ is $n_s$ and the dimensionality of the angular features $\mathbf{F}_a$ is $n_a$. We denote a training configuration with a triplet $(n_s, n_a, l)$. A \cmark-symbol denotes the training of per-camera bias vectors, an \xmark-symbol is used otherwise.

\section{Neural Architecture comparison}

Table \ref{tab:neural_architecture} gives a comparison of several neural architectures used in the context of implicit surface reconstruction. Thanks to our light field probes, our MLP is entirely agnostic to the surface orientation and position, hence we can reduce its size and obtain high quality renderings, in real-time.

\begin{table}[htb]
	\begin{center}
	\scriptsize
	\begin{tabular}{|c|c|c|c|}
	\hline
    Method  & Grid Type & \makecell{SDF MLP \\ Layers / Neurons} & \makecell{Color MLP \\ Layers / Neurons} \\
	\hline
	NeuS & \xmark & 8 / 256 & 4 / 256 \\
	\hline
	NeuS2 & hash-grid & 1 / 64 & 2 / 64 \\
	\hline
	Voxurf & dense & \xmark & 4 / 192 \\
	\hline
	Ours & sparse & \xmark & 2 / 32 \\
	\hline
	\end{tabular}
	\caption{Neural architectures in the literature.}
	\label{tab:neural_architecture}
	\end{center}
\end{table}

\section{Comparisons on ActorsHQ}

\begin{figure}[htb]
\includegraphics[width=\linewidth]{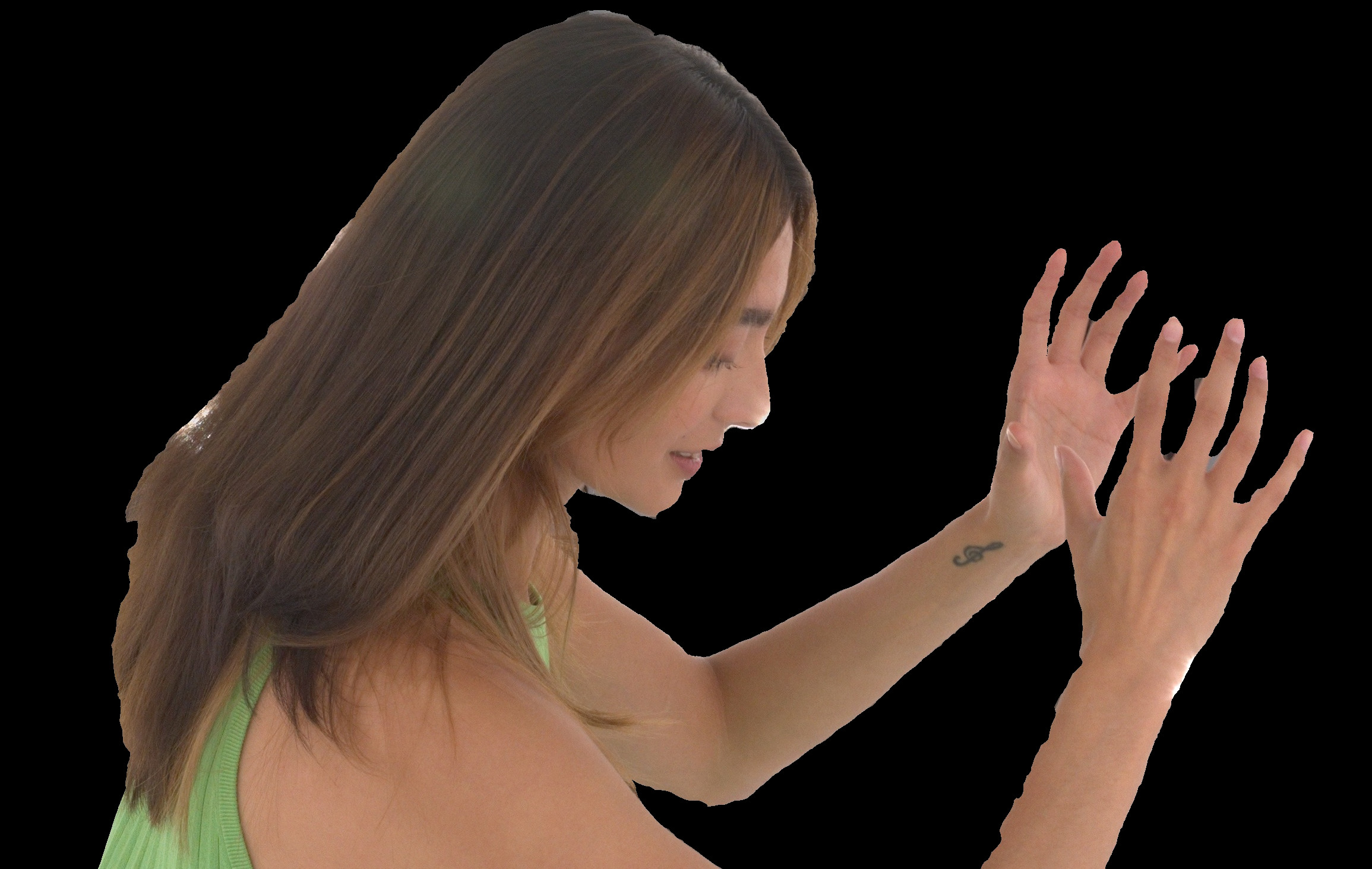}
\caption{Imprecise segmentation example.}
\label{fig:ActorsHQSegmentation}
\end{figure}

\begin{figure*}[h]
\includegraphics[width=\textwidth]{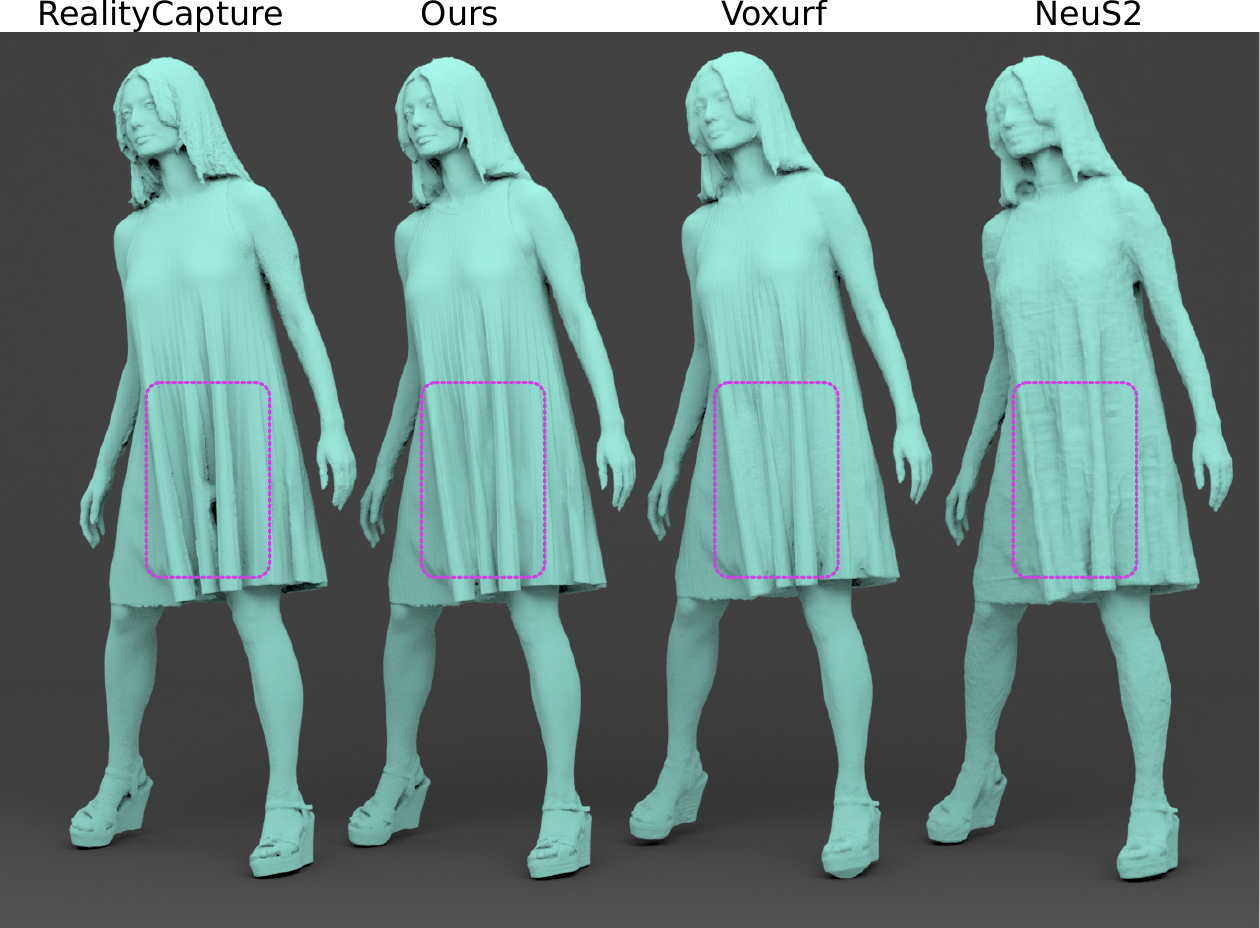}
\caption{Reconstruction results on ActorsHQ. Left to right: RealityCapture, Ours, Voxurf, NeuS2.}
\label{fig:actorshq_geom_comp1}
\end{figure*}

\begin{figure*}[h]
\includegraphics[width=\textwidth]{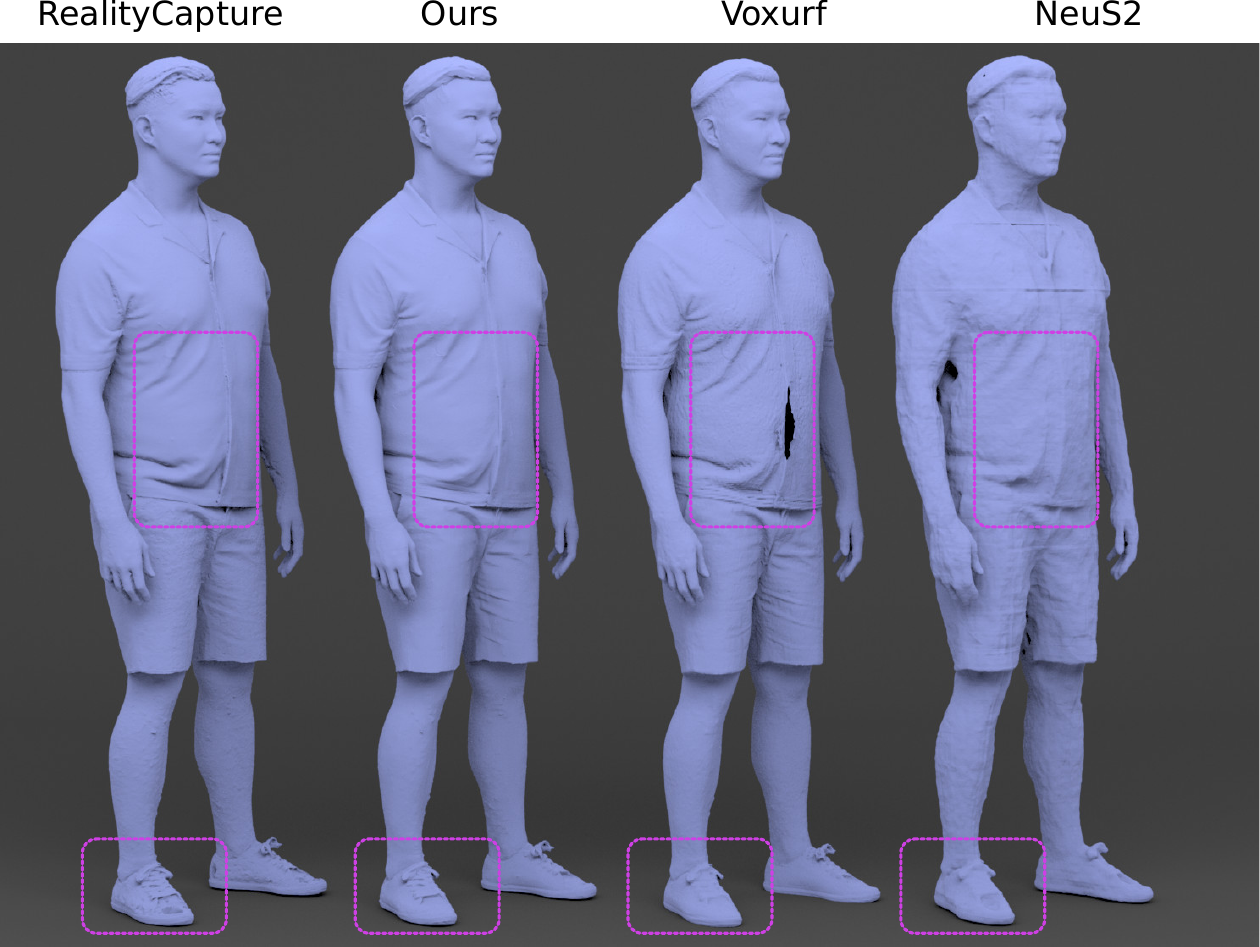}
\caption{Reconstruction results on ActorsHQ. Left to right: RealityCapture, Ours, Voxurf, NeuS2.}
\label{fig:actorshq_geom_comp2}
\end{figure*}

\begin{figure*}[h]
\includegraphics[width=\textwidth]{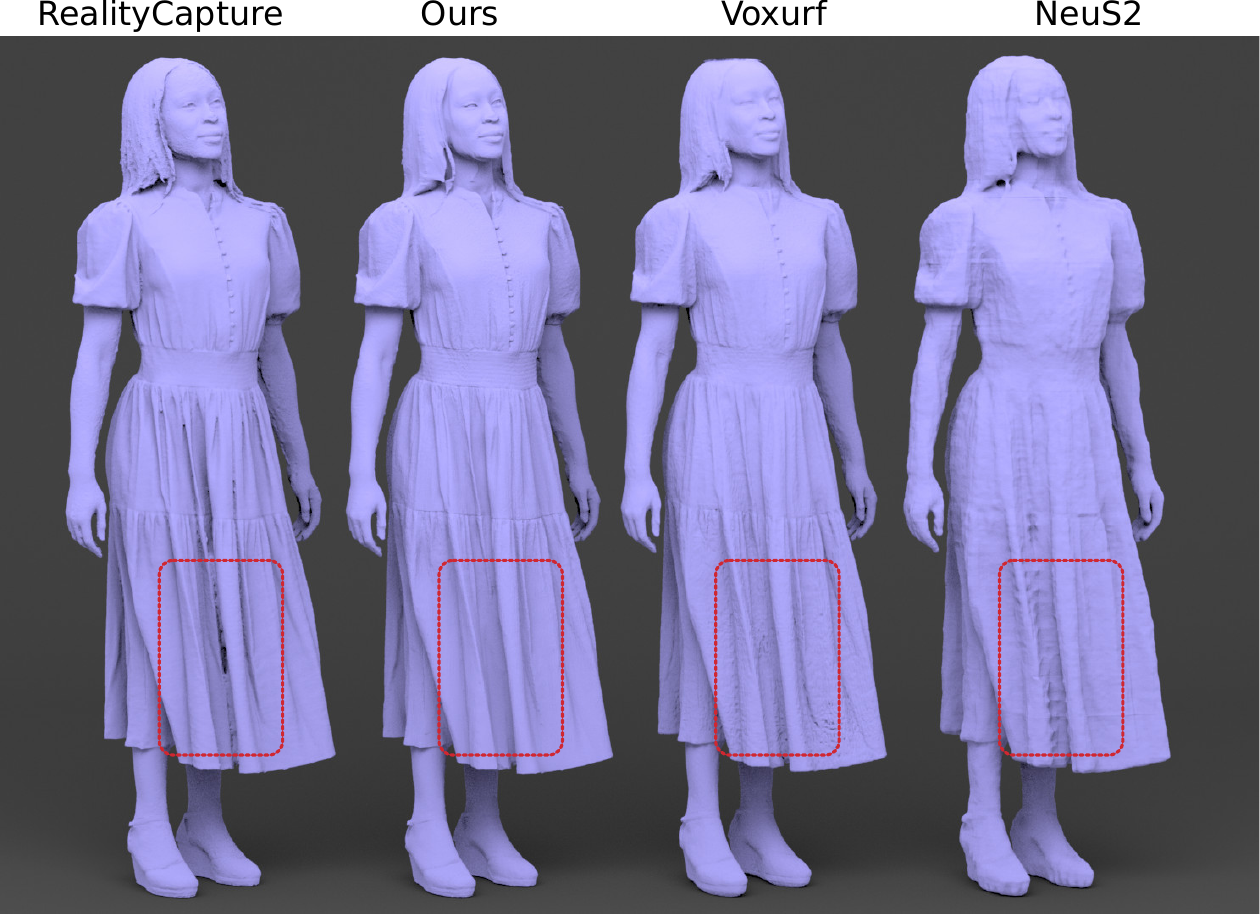}
\caption{Reconstruction results on ActorsHQ. Left to right: RealityCapture, Ours, Voxurf, NeuS2.}
\label{fig:actorshq_geom_comp3}
\end{figure*}

\begin{figure*}[h]
\includegraphics[width=\textwidth]{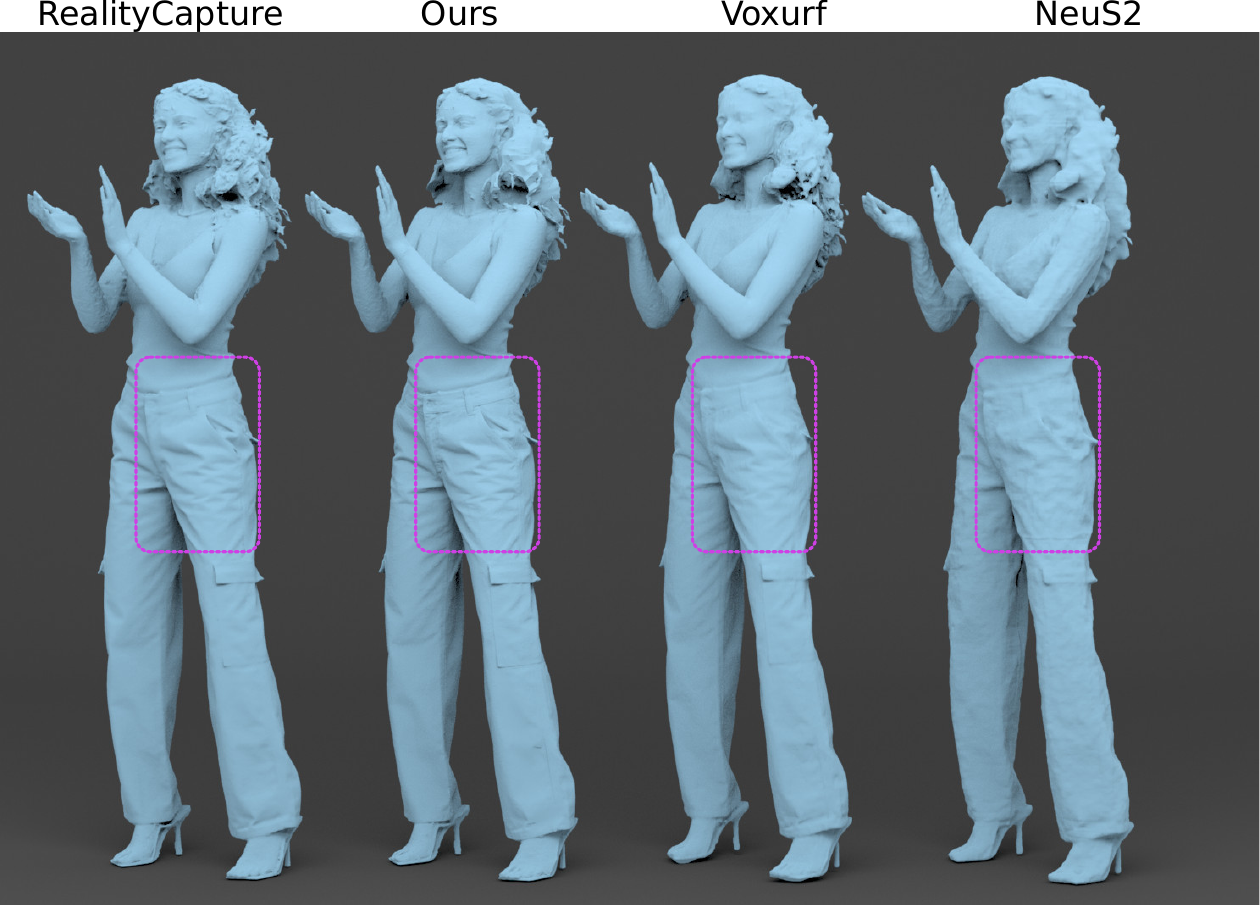}
\caption{Reconstruction results on ActorsHQ. Left to right: RealityCapture, Ours, Voxurf, NeuS2.}
\label{fig:actorshq_geom_comp4}
\end{figure*}

\begin{figure*}[h]
\includegraphics[width=\textwidth]{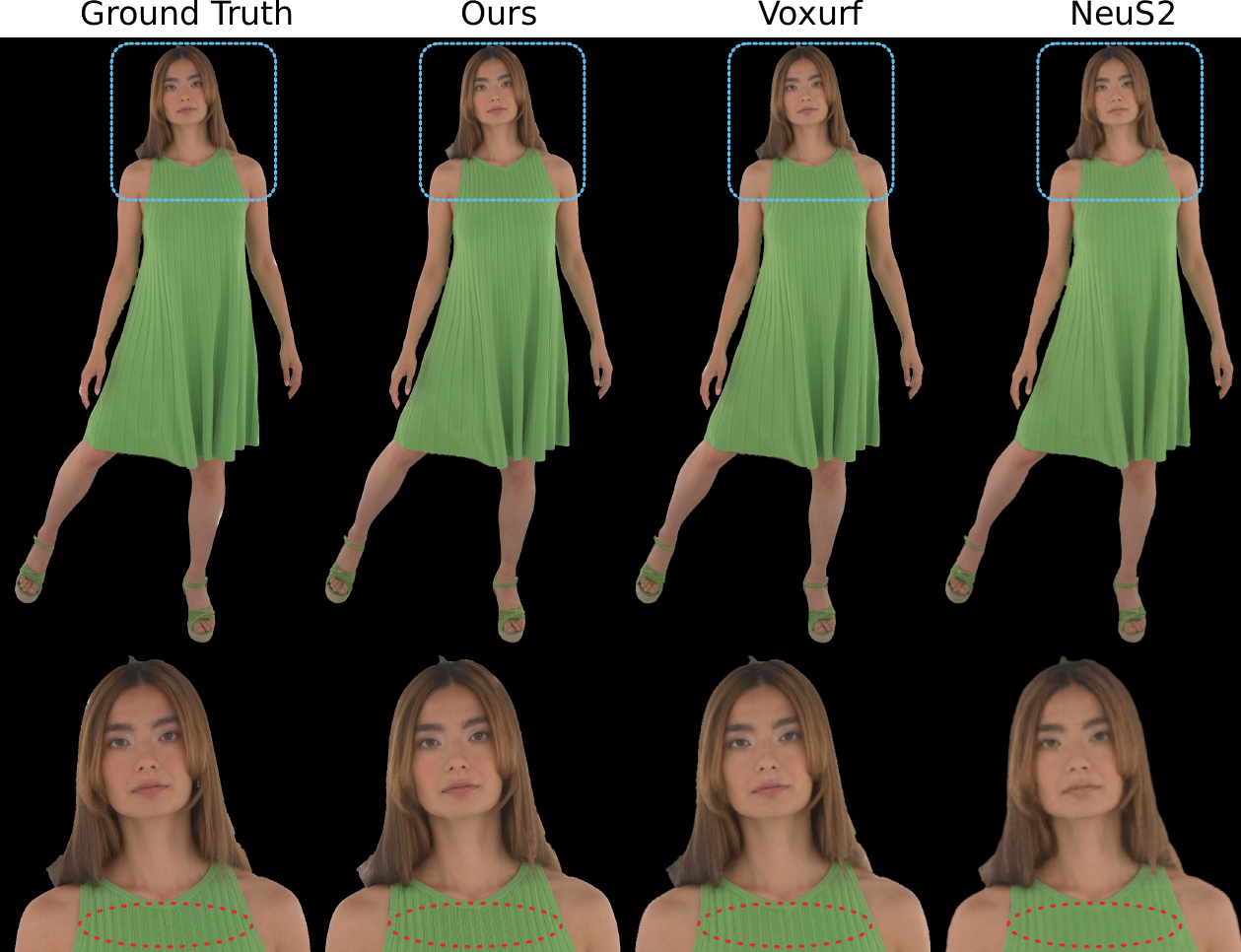}
\caption{Qualitative comparison on ActorsHQ. Left to right: ground truth, Ours, Voxurf, NeuS2. Our method is able to handle the full resolution images, which enables to reconstruct the sewing patterns at a sub-millimetric scale.}
\label{fig:actorshq_qualitative_comp1}
\end{figure*}

\begin{figure*}[h]
\includegraphics[width=\textwidth]{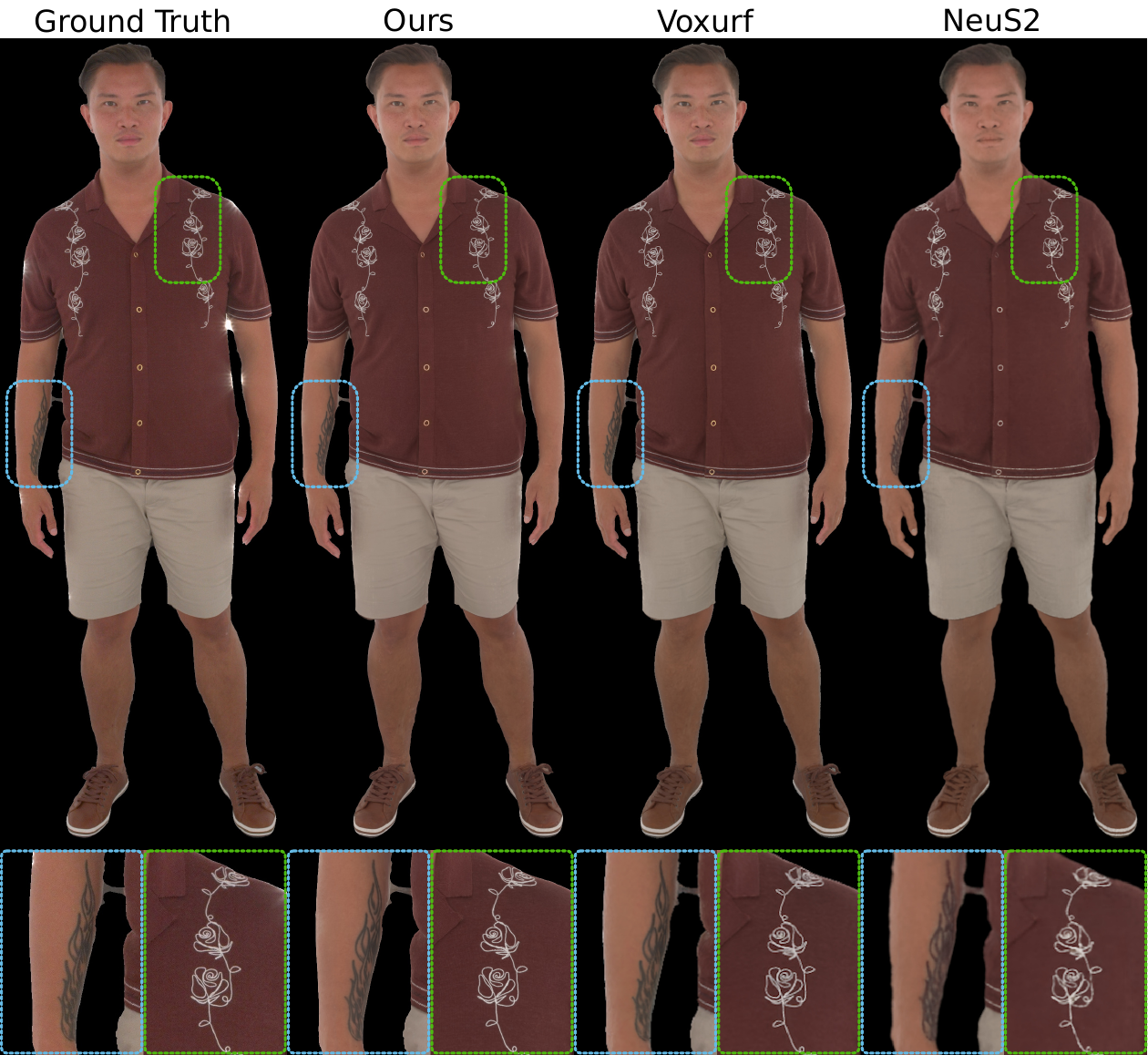}
\caption{Qualitative comparison on ActorsHQ. Left to right: ground truth, Ours, Voxurf, NeuS2.}
\label{fig:actorshq_qualitative_comp2}
\end{figure*}

\begin{figure*}[h]
\includegraphics[width=\textwidth]{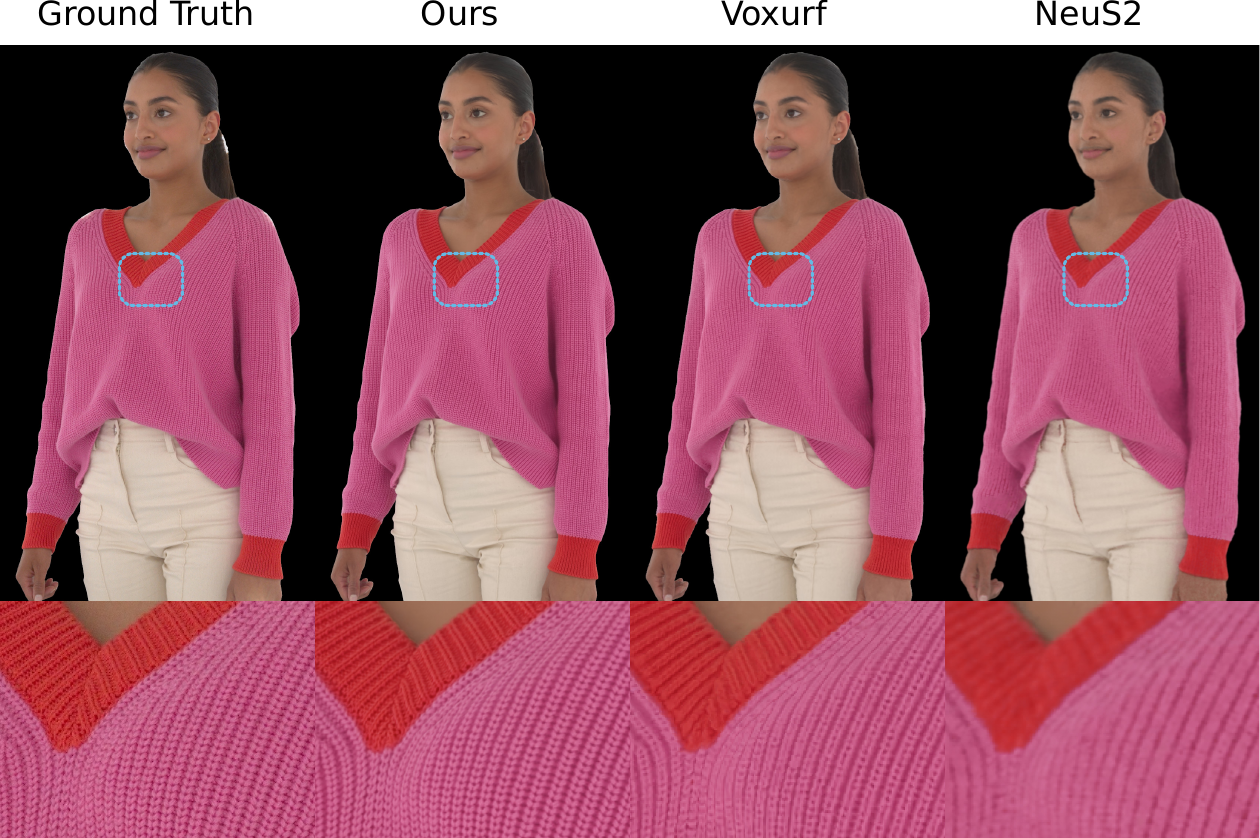}
\caption{Qualitative comparison on ActorsHQ. Left to right: ground truth, Ours, Voxurf, NeuS2.}
\label{fig:actorshq_qualitative_comp3}
\end{figure*}

Geometric comparisons on the ActorsHQ dataset \cite{isik2023humanrf} are shown in figures \ref{fig:actorshq_geom_comp1}, \ref{fig:actorshq_geom_comp2}, \ref{fig:actorshq_geom_comp3}, \ref{fig:actorshq_geom_comp4}. The dataset comes with meshes reconstructed by RealityCapture \cite{RealityCapture}, a multi-view stereo reconstruction software. We cannot compute geometric metrics since there is no ground truth obtained independently from the images. A qualitative comparison of the volume rendering quality is shown in figures \ref{fig:actorshq_qualitative_comp1}, \ref{fig:actorshq_qualitative_comp2} and \ref{fig:actorshq_qualitative_comp3}. We used the (4,4,4) configuration here. Note that the input images come with pre-baked segmentation masks, as shown in figure \ref{fig:ActorsHQSegmentation}, that tend to have poorer accuracy on the arms and hands. This results in both geometric and photometric artifacts that are difficult to eliminate.



\section{Comparisons on DTU}

\begin{figure*}[h]
\includegraphics[width=\textwidth]{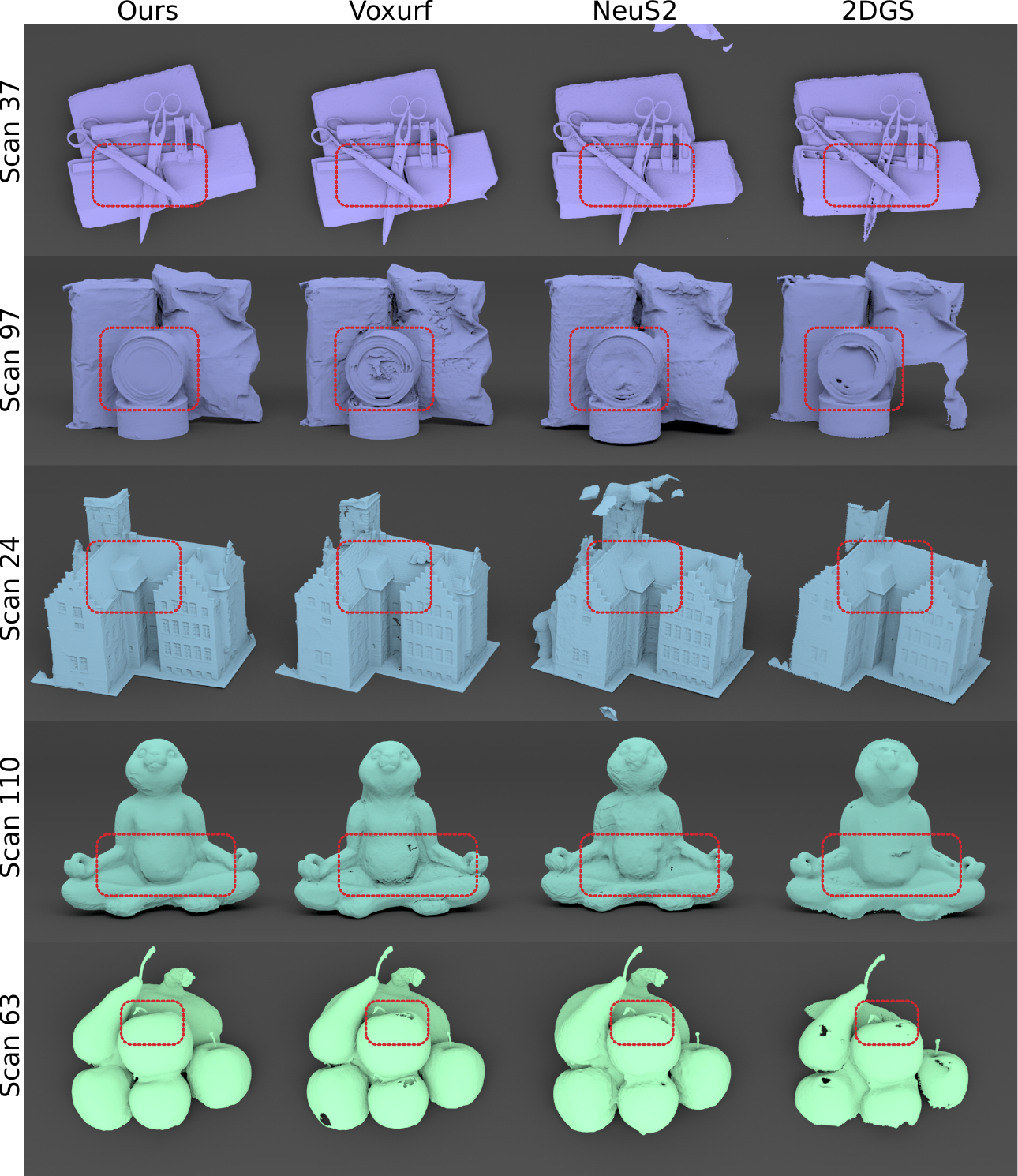}
\caption{Reconstruction results on DTU. Left to right: Ours, Voxurf, NeuS2, 2DGS. We find that 2DGS excels at reconstructing flat surfaces (doll house roof) but tends to under-perform on reflective materials. 2DGS fails to extract geometry on some parts of the objects (scans 97 and 63). In contrast, our method recovers smooth surfaces even under strong specularities (metal scissors, tuna can and apples). Voxurf struggles on the most shiny materials despite its considerably larger MLP. NeuS2's reconstruction suffers from grid-aligned artifacts, possibly due to discontinuities in its hash-grid interpolation scheme (shoulder of the bunny in scan 110).}
\label{fig:dtu_meshes}
\end{figure*}

\begin{figure*}[h]
\includegraphics[width=1.0\textwidth]{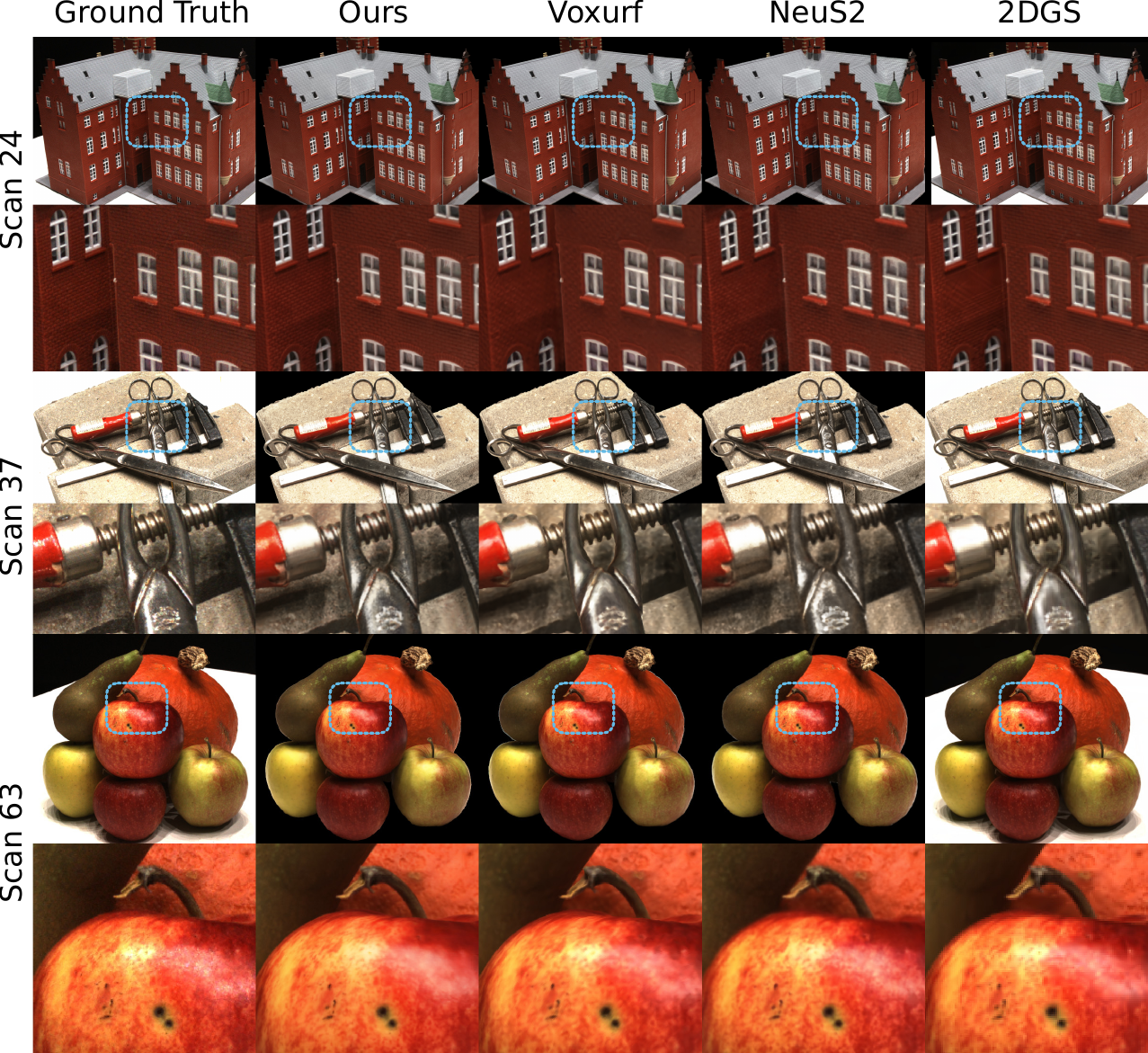}
\caption{Qualitative comparison on DTU. Top: scan 24, middle: scan 37, bottom: scan 63. Left to right: ground truth, Ours, Voxurf, NeuS2, 2DGS.}
\label{fig:dtu_qualitative_comp1}
\centering
\end{figure*}

\begin{figure*}[h]
\includegraphics[width=1.0\textwidth]{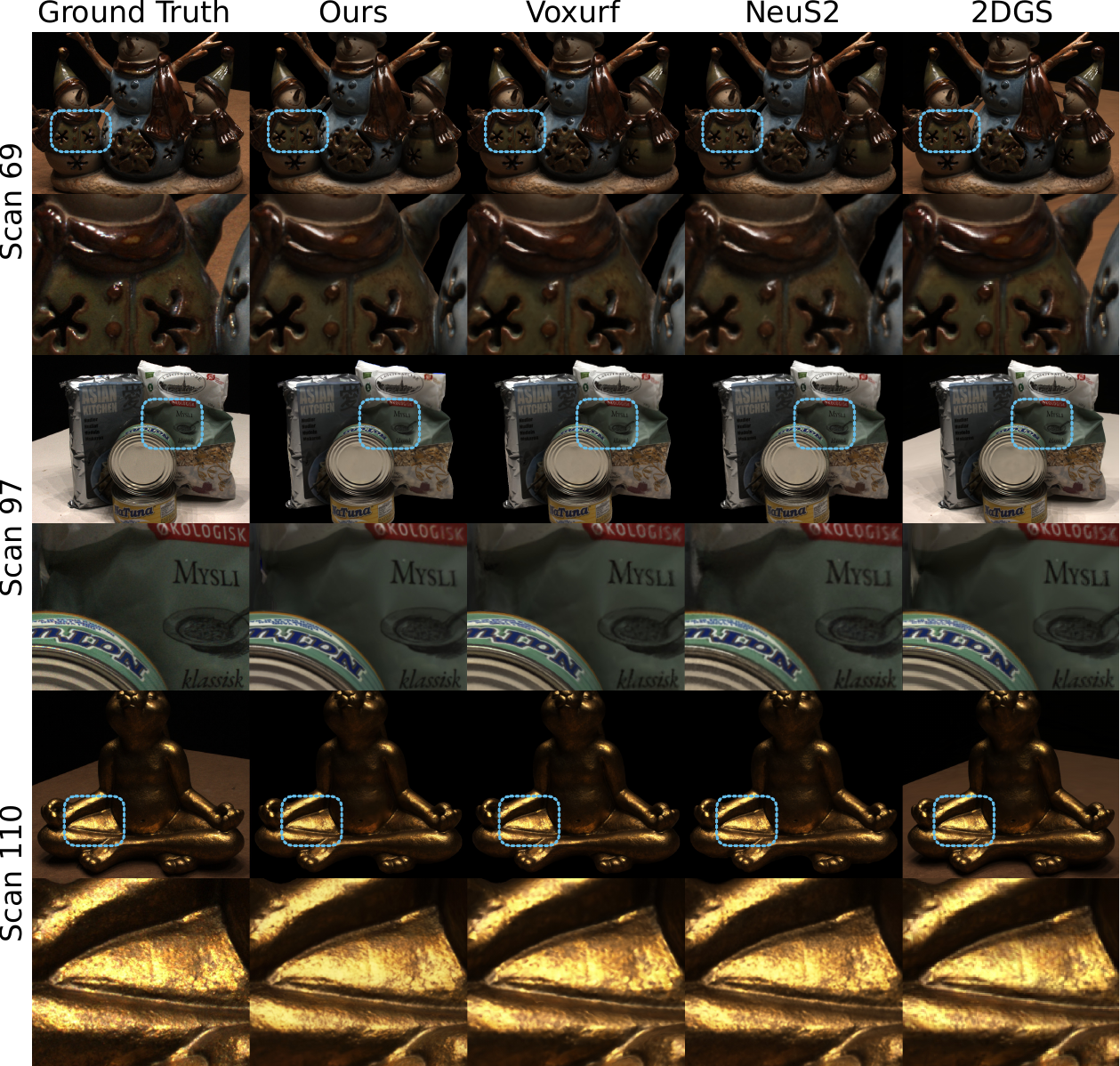}
\caption{Qualitative comparison on DTU. Top: scan 69, middle: scan 97, bottom: scan 110. Left to right: ground truth, Ours, Voxurf, NeuS2, 2DGS.}
\label{fig:dtu_qualitative_comp2}
\centering
\end{figure*}

Figure \ref{fig:dtu_meshes} presents a comparison of some of the reconstruction results on DTU \cite{jensen2014large}. Close-ups of the volume rendered images are shown in figures \ref{fig:dtu_qualitative_comp1} and \ref{fig:dtu_qualitative_comp2}. Our results are obtained with the (4,4,4) configuration.

\section{Comparisons on BlendedMVS}

\begin{figure*}[h]
\includegraphics[width=\textwidth]{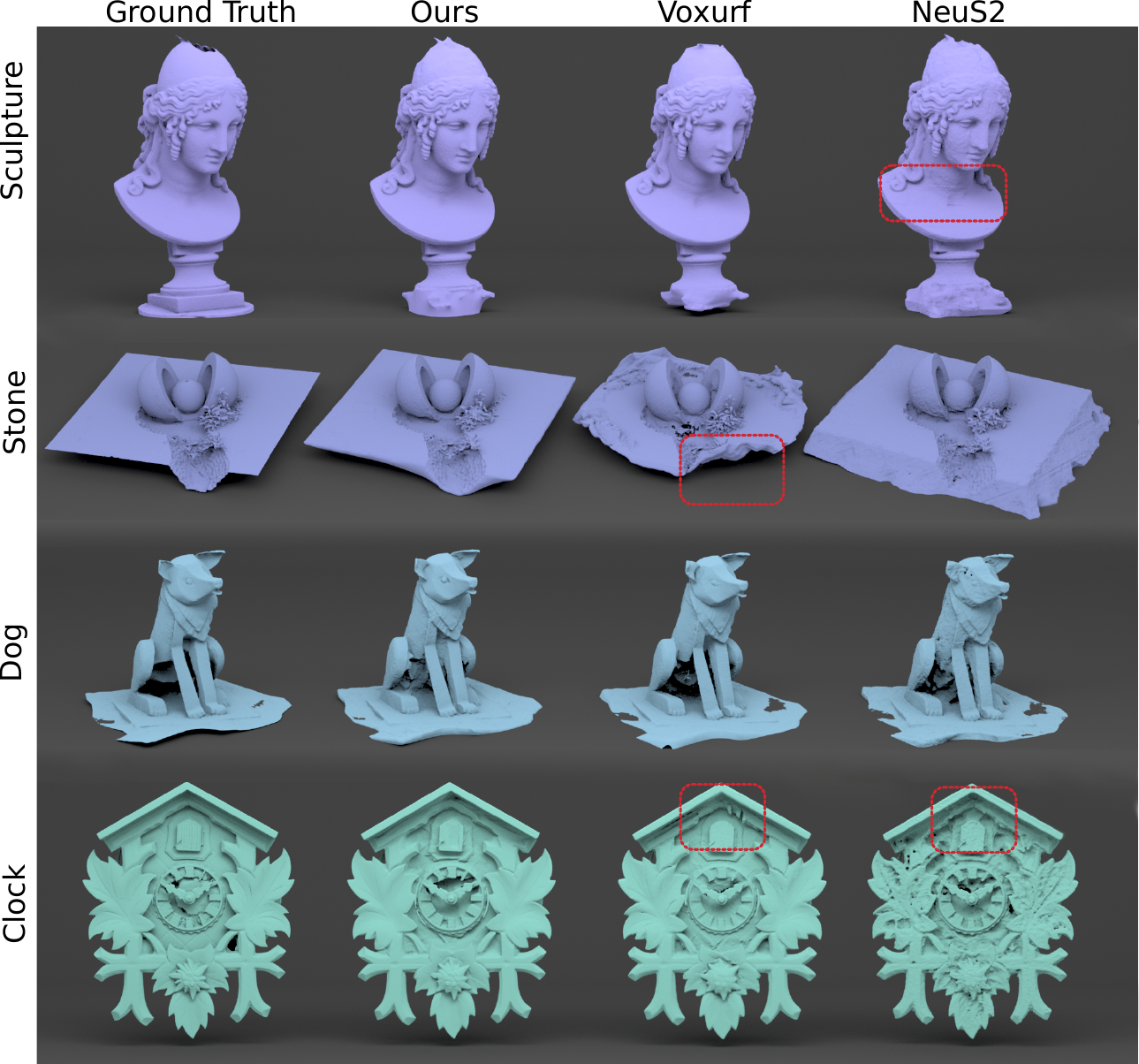}
\caption{Reconstruction results on BlendedMVS. Left to right: Ground Truth, Ours, Voxurf, NeuS2.}
\label{fig:bmvs_geom_comp_suppmat}
\centering
\end{figure*}

\begin{figure*}[h]
\includegraphics[width=0.9\textwidth]{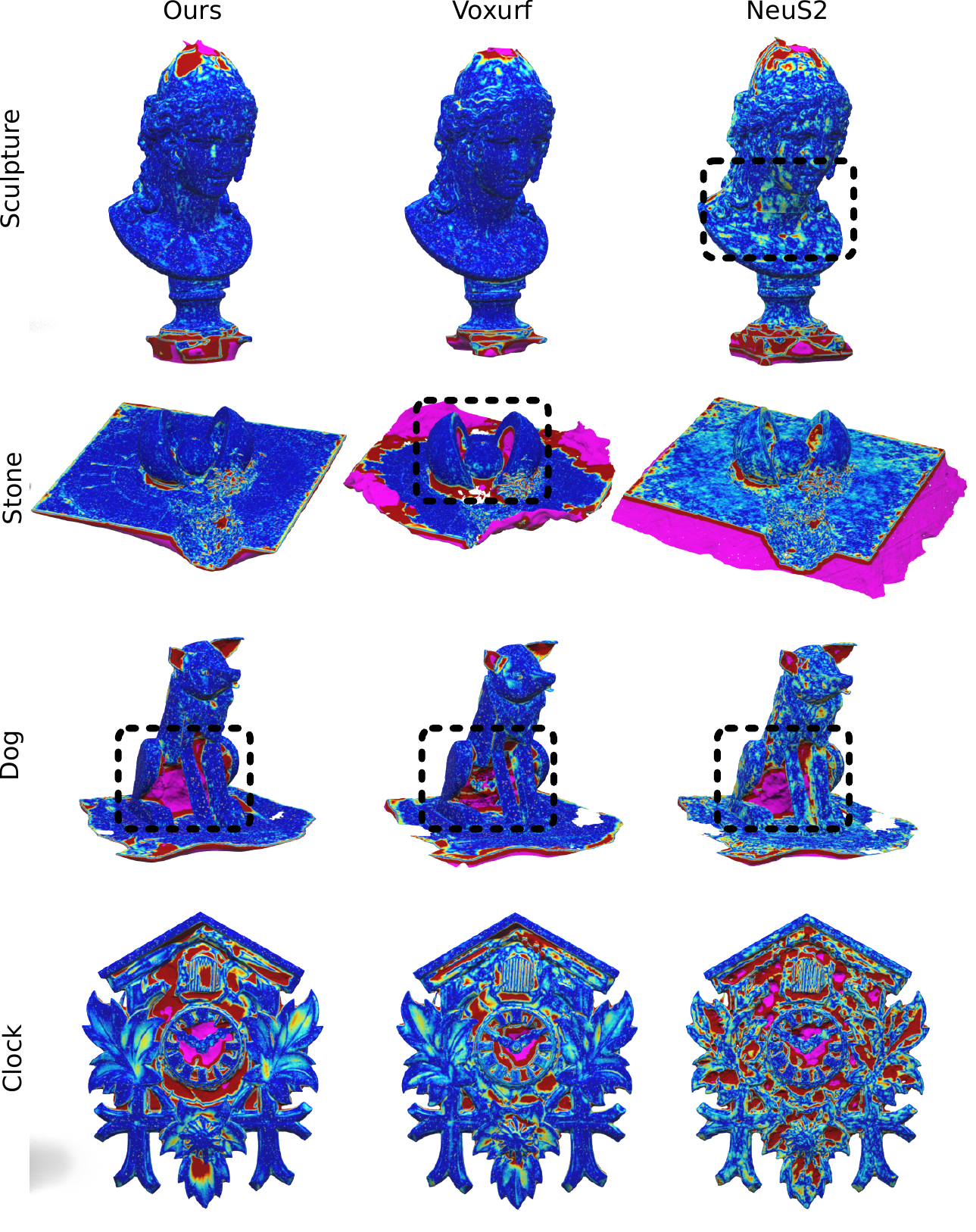}
\caption{Accuracy heatmaps on BlendedMVS. The pink color denotes points too far away from the ground truth, which are ignored in the computation of the metrics. Left to right: Ours, Voxurf, NeuS2. Voxurf fails to carve inside the two hemispheres in the stone example and the corners of the base are missing. However, Voxurf is able to carve under the dog statue whereas both NeuS2 and our method fail on this example. NeuS2 is noticably more noisy on all examples.}
\label{fig:bmvs_acc_comp}
\centering
\end{figure*}

\begin{figure*}[h]
\centering
\includegraphics[width=0.9\textwidth]{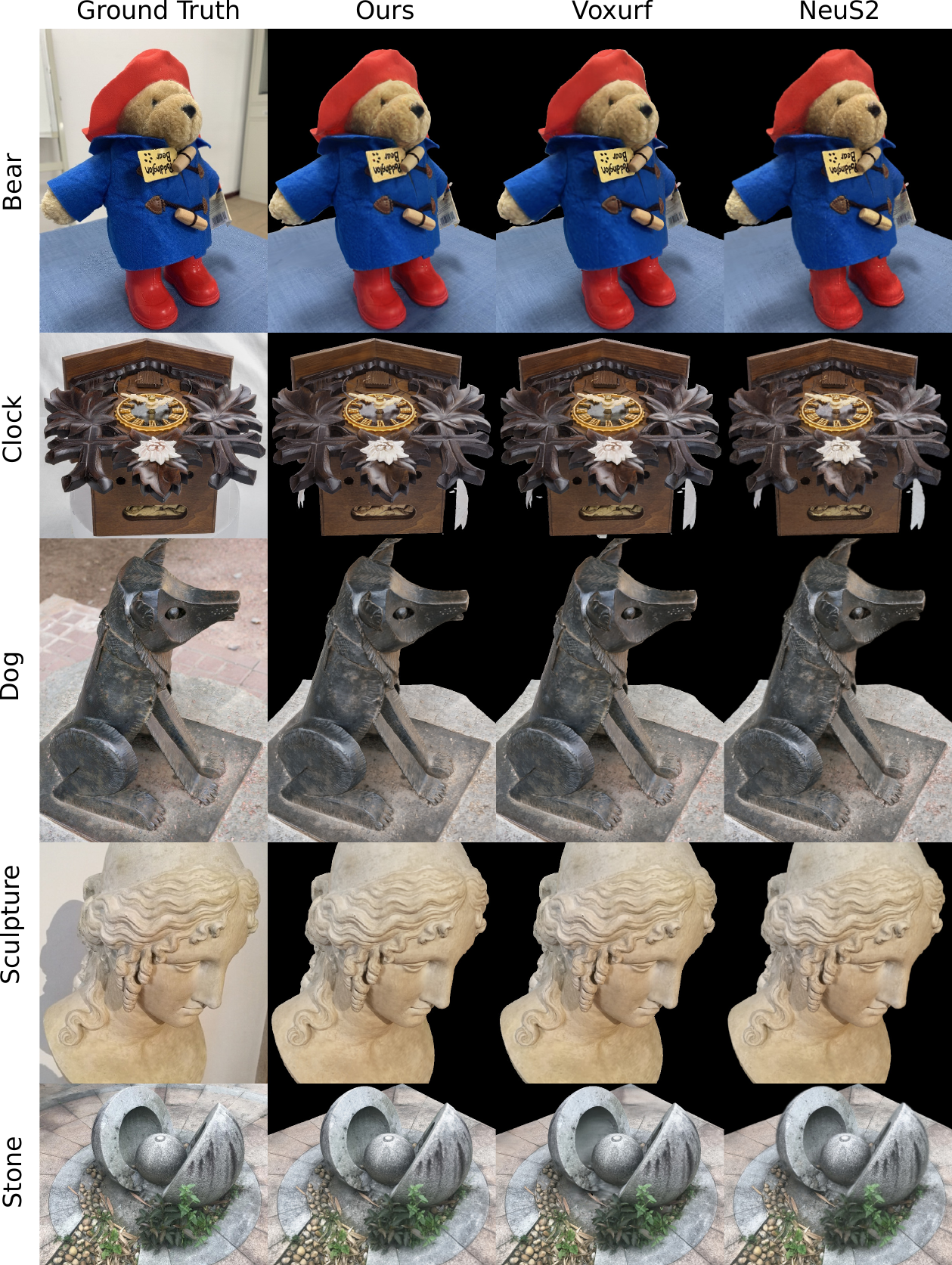}
\caption{Qualitative comparison on BlendedMVS. Left to right: ground truth, Ours, Voxurf, NeuS2.}
\label{fig:bmvs_qualitative_comp1}
\centering
\end{figure*}

Geometric comparisons on the BlendedMVS dataset \cite{yao2020blendedmvs} are shown in figures \ref{fig:bmvs_geom_comp_suppmat} and \ref{fig:bmvs_acc_comp}. Qualitative comparisons of the volume rendering are shown in figure \ref{fig:bmvs_qualitative_comp1}. We used the (8,8,4) configuration here as it performed a little better on this dataset (see table \ref{tab:bmvs_chamfer}).

\section{Performance Analysis}

We record inference timings on the apples example (scan 63 of DTU, (4,4,4) configuration), with a window of resolution 1920$\times$1163 and present the results in table \ref{tab:inference_perf}. We lock the memory clock to 5001MHz and the gpu clock to 1500MHz to obtain stable performance measurements. The  shading column corresponds to the assignment of a color to each voxel. The render column corresponds to the volume rendering kernel, which samples the SDF and color fields along rays to generate the final image. 
The first 4 rows correspond to the fully-fused color prediction kernel, with the computation of the spatial and angular features enabled or disabled. Thus, the 4th row corresponds to the MLP inference only whereas the 1st row corresponds to the full model. The last two rows correspond to the computation of $\mathbf{F}_s$ or $\mathbf{F}_a$ on their own, in separate kernels, and whose result is interpreted as a per-voxel color for visualization. 
\begin{table}[htb]
	\begin{center}
	\small
	\begin{tabular}{|c|c|c|}
	\hline
    Type     & Shading & Render \\
	\hline
	\hline
    MLP \cmark, $\mathbf{F}_s$ \cmark, $\mathbf{F}_a$ \cmark &    2.47   &  1.34 \\
	\hline
    MLP \cmark, $\mathbf{F}_s$ \cmark, $\mathbf{F}_a$ \xmark & 1.92  & 1.34 \\
	\hline
    MLP \cmark, $\mathbf{F}_s$ \xmark, $\mathbf{F}_a$ \cmark & 2.32  & 1.34 \\
	\hline
    MLP \cmark, $\mathbf{F}_s$ \xmark, $\mathbf{F}_a$ \xmark  & 1.80 & 1.34 \\
	\hline
    \hline
    $\mathbf{F}_a$ only   &   1.96  &   1.34 \\
	\hline
    $\mathbf{F}_s$ only  &    0.82  &   1.34 \\
	\hline
    		
	\end{tabular}
	\caption{Timings in ms}
	\label{tab:inference_perf}
	\end{center}
\end{table}

We observe that just evaluating the MLP (4th row) or computing the angular features on their own (5th row) roughly takes the same amount of time (1.80ms and 1.96ms) but that fusing the two operations together only takes 2.32ms, much less than the sum of the two ($1.80+1.96=3.76$ms). Including the computation of the spatial features gives the full model at 2.47ms. 


\begin{table*}[htb]
	\begin{center}
	\scriptsize
		\begin{tabular}{|c||c||c|c|c|c|c|c|c|c|c|c|c|c|c|c|}
			\hline
			&  & \multicolumn{8}{c|}{kinette} & \multicolumn{6}{c|}{kino} \\
			\hline
			Chamfer (mm) & Mean & cos & naked & jea & opt1 & opt2 & opt3 & sho & tig & cos & naked & jea & opt & sho & tig \\
			\hline
			(4,4,4)\xmark & \textbf{1.04} & \textbf{1.51} & \textbf{0.54} & \textbf{1.09} & \textbf{1.14} & \textbf{1.75} & \textbf{1.22} & \textbf{0.95} & \textbf{0.66} & \textbf{1.53} & \underline{0.60} & \textbf{0.75} & \textbf{1.51} & \textbf{0.67} & \textbf{0.67} \\
			\hline
			MMH & \underline{1.15} & \underline{1.55} & \textbf{0.54} & \underline{1.10} & \underline{1.16} & 2.12 & \underline{1.45} & \underline{1.07} & \underline{0.68} & \underline{1.77} & \textbf{0.59} & \underline{0.80} & \underline{1.72} & \underline{0.75} & \underline{0.75} \\
			\hline
			Voxurf & 1.59 & 1.70 & \underline{1.28} & 1.62 & 1.62 & \underline{2.06} & 1.61 & 1.48 & 0.98 & 2.60 & 1.30 & 1.27 & 2.54 & 1.12 & 1.06 \\
			\hline
			Neus2 & 2.13 & 3.71 & 1.43 & 2.10 & 2.13 & 2.90 & 2 & 2.12 & 1.25 & 3.55 & 1.61 & 1.64 & 2.61 & 1.43 & 1.37 \\
			\hline
			Colmap & 3.51 & 2.69 & 4.27 & 2.99 & 4.59 & 4.18 & 2.81 & 3.71 & 2.50 & 4.11 & 4.20 & 2.48 & 4.34 & 3.09 & 3.23 \\
			\hline
			2DGS & 3.35 & 5.05 & 2.27 & 8.30 & 3.25 & 3.40 & 3.63 & 2.66 & 3.52 & 4.18 & 2.25 & 1.75 & 2.80 & 1.98 & 1.87 \\
			\hline
		\end{tabular}
	\caption{MVMannequins per-scene chamfer (mm)}
	\label{tab:mvmannequins_chamfer}
	\end{center}
	\begin{center}
	\scriptsize
		\begin{tabular}{|c||c||c|c|c|c|c|c|c|c|c|c|c|c|c|c|}
			\hline
			 &  & \multicolumn{8}{c|}{kinette} & \multicolumn{6}{c|}{kino} \\
			\hline
			PSNR (db) & Mean & cos & naked & jea & opt1 & opt2 & opt3 & sho & tig & cos & naked & jea & opt & sho & tig \\
			\hline
			(4,4,4)\xmark & \textbf{36.81} & \textbf{29.48} & \textbf{40.61} & \textbf{30.73} & \textbf{40.48} & \textbf{38.69} & \textbf{34.13} & \textbf{37.70} & \textbf{31.74} & \textbf{39.64} & \textbf{41.57} & \textbf{35.80} & \textbf{39.29} & \textbf{37.50} & \textbf{37.99} \\
			\hline
			MMH & \underline{36.33} & \underline{29.24} & \underline{40.03} & \underline{30.55} & \underline{39.94} & \underline{38.01} & \underline{33.82} & \underline{37.18} & \underline{31.38} & \underline{39.22} & \underline{40.88} & \underline{35.25} & \underline{38.59} & \underline{37.11} & \underline{37.49} \\
			\hline
			Voxurf & 35.51 & 28.39 & 37.51 & 30.20 & 39.26 & 37.34 & 32.82 & 36.73 & 30.81 & 38.17 & 40.09 & 34.45 & 37.57 & 36.69 & 37.16 \\
			\hline
			Neus2 & 34.22 & 28.09 & 37.23 & 29.44 & 36.98 & 35.55 & 32.58 & 34.98 & 30.14 & 36.65 & 38.05 & 33.39 & 35.78 & 35.01 & 35.23 \\
			\hline
			2DGS & 34.89 & 27.26 & 37.92 & 28.88 & 37.62 & 36.84 & 32.22 & 36.24 & 29.86 & 37.97 & 38.82 & 34.99 & 37.28 & 36.14 & 36.50 \\
			\hline
		\end{tabular}
	\caption{MVMannequins per-scene PSNR}
	\end{center}

	\begin{center}
	\scriptsize
		\begin{tabular}{|c||c||c|c|c|c|c|c|c|c|c|c|c|c|c|c|}
			\hline
			 & & \multicolumn{8}{c|}{kinette} & \multicolumn{6}{c|}{kino} \\
			\hline
			Chamfer (mm) & Mean & cos & naked & jea & opt1 & opt2 & opt3 & sho & tig & cos & naked & jea & opt & sho & tig \\
			\hline
			(8,8,4)\xmark & \textbf{1.03} & \textbf{1.48} & \textbf{0.54} & \textbf{1.08} & \underline{1.12} & \textbf{1.70} & 1.24 & \textbf{0.93} & \textbf{0.65} & 1.52 & \textbf{0.59} & 0.74 & 1.51 & \textbf{0.66} & \underline{0.67} \\
			\hline
			(12,12,4)\xmark & 1.04 & 1.52 & \underline{0.54} & 1.09 & 1.13 & 1.72 & 1.23 & 0.96 & 0.66 & \underline{1.52} & \underline{0.59} & \textbf{0.74} & 1.52 & 0.68 & \textbf{0.67} \\
			\hline
			(4,4,1)\xmark & 1.30 & 1.56 & 1.22 & 1.22 & 1.25 & 1.78 & 1.62 & 1.50 & 1.05 & 1.78 & 1.16 & 0.81 & \underline{1.48} & 0.86 & 0.95 \\
			\hline
			(4,4,2)\xmark & \underline{1.04} & \underline{1.48} & 0.59 & 1.09 & \textbf{1.10} & \underline{1.71} & \textbf{1.22} & \underline{0.94} & 0.67 & \textbf{1.51} & 0.64 & \underline{0.74} & \textbf{1.45} & 0.69 & 0.69 \\
			\hline
			(4,4,3)\xmark & 1.04 & 1.49 & 0.56 & \underline{1.09} & 1.12 & 1.73 & 1.22 & 0.96 & 0.66 & 1.52 & 0.60 & 0.75 & 1.50 & 0.68 & 0.68 \\
			\hline
			(4,4,4)\xmark & 1.04 & 1.51 & 0.54 & 1.09 & 1.14 & 1.75 & \underline{1.22} & 0.95 & \underline{0.66} & 1.53 & 0.60 & 0.75 & 1.51 & \underline{0.67} & 0.67 \\
			\hline
		\end{tabular}
	\caption{Detailed Ablation Table. MVMannequins per-scene chamfer (mm)}
	\end{center}

	\begin{center}
	\scriptsize
		\begin{tabular}{|c||c||c|c|c|c|c|c|c|c|c|c|c|c|c|c|}
			\hline
			 & & \multicolumn{8}{c|}{kinette} & \multicolumn{6}{c|}{kino} \\
			\hline
			PSNR (db) & Mean & cos & naked & jea & opt1 & opt2 & opt3 & sho & tig & cos & naked & jea & opt & sho & tig \\
			\hline
			(8,8,4)\xmark & \underline{36.90} & \underline{29.56} & \underline{40.76} & \textbf{30.80} & \underline{40.62} & \underline{38.80} & \underline{34.23} & \underline{37.76} & \textbf{31.86} & \underline{39.76} & \underline{41.66} & \underline{35.88} & \underline{39.35} & \underline{37.56} & \underline{38.07} \\
			\hline
			(12,12,4)\xmark & \textbf{36.93} & \textbf{29.57} & \textbf{40.79} & \underline{30.79} & \textbf{40.63} & \textbf{38.82} & \textbf{34.24} & \textbf{37.78} & \underline{31.83} & \textbf{39.79} & \textbf{41.73} & \textbf{35.88} & \textbf{39.41} & \textbf{37.58} & \textbf{38.20} \\
			\hline
			(4,4,1)\xmark & 35.84 & 29.24 & 38.61 & 30.52 & 39.03 & 37.47 & 33.55 & 36.24 & 31.24 & 38.81 & 39.46 & 35.37 & 38.68 & 36.58 & 36.98 \\
			\hline
			(4,4,2)\xmark & 36.53 & 29.31 & 40.09 & 30.62 & 40.07 & 38.38 & 33.95 & 37.42 & 31.55 & 39.35 & 41.06 & 35.67 & 39.01 & 37.26 & 37.71 \\
			\hline
			(4,4,3)\xmark & 36.68 & 29.38 & 40.39 & 30.66 & 40.30 & 38.54 & 34.05 & 37.56 & 31.62 & 39.50 & 41.37 & 35.72 & 39.14 & 37.41 & 37.86 \\
			\hline
			(4,4,4)\xmark & 36.81 & 29.48 & 40.61 & 30.73 & 40.48 & 38.69 & 34.13 & 37.70 & 31.74 & 39.64 & 41.57 & 35.80 & 39.29 & 37.50 & 37.99 \\
			\hline
		\end{tabular}
	\caption{Detailed Ablation Table. MVMannequins per-scene PSNR}
	\end{center}
\end{table*}


\begin{table*}[htb]
	\begin{center}
	\scriptsize
		\begin{tabular}{|c|c||c||c|c|c|c|c|c|c|c|c|c|c|c|c|}
			\hline
			PSNR & Resolution & Mean & A1S1 & A2S1 & A3S1 & A4S1 & A5S1 & A6S1 & A7S1 & A8S1 & A1S2 & A4S2 & A5S2 & A6S2 & A8S2 \\
			\hline
			(4,4,4)\xmark & r/1 & \textbf{37.48} & \textbf{37.64} & \textbf{38.21} & \textbf{37.58} & \textbf{35.78} & \textbf{38.36} & \underline{36.80} & \textbf{37.80} & \textbf{38.09} & \textbf{37.91} & \textbf{35.59} & \textbf{38.56} & \underline{36.59} & \textbf{38.32} \\
			\hline
			(4,4,4)\xmark & r/2 & \underline{36.62} & 36.86 & 36.90 & \underline{36.59} & 34.55 & \underline{37.54} & 36.39 & \underline{37.03} & \underline{37.28} & \underline{37.19} & 34.32 & \underline{37.75} & 36.24 & \underline{37.47} \\
			\hline
			(4,4,4)\xmark & r/4 & 34.75 & 35.43 & 34.05 & 35.21 & 32.08 & 35.59 & 34.96 & 35.41 & 35.16 & 35.67 & 32.14 & 35.94 & 34.75 & 35.34 \\
			\hline
			Voxurf & r/2 & 36.56 & \underline{37.04} & \underline{36.93} & 36.17 & \underline{34.58} & 36.96 & \textbf{36.85} & 36.69 & 36.99 & 37.12 & \underline{34.33} & 37.53 & \textbf{36.81} & 37.31 \\
			\hline
			Neus2 & r/2 & 34.53 & 35.22 & 34.67 & 33.28 & 32.50 & 35.20 & 33.86 & 35.43 & 35.41 & 35.71 & 32.52 & 35.38 & 34.02 & 35.72 \\
			\hline
		\end{tabular}
	\caption{ActorsHQ per-scene PSNR}
	\label{tab:actorshq_table}
	\end{center}
\end{table*}


\begin{table*}[htb]
	\scriptsize
	\begin{center}
		\begin{tabular}{|c||c||c|c|c|c|c|c|c|c|c|c|c|c|c|c|c|}
			\hline
			Chamfer (mm) & Mean & 24 & 37 & 40 & 55 & 63 & 65 & 69 & 83 & 97 & 105 & 106 & 110 & 114 & 118 & 122 \\
			\hline
			(4,4,4)\cmark & \textbf{0.68} & 0.65 & 0.74 & \underline{0.34} & \textbf{0.34} & 1.02 & \underline{0.71} & \textbf{0.62} & 1.34 & \textbf{0.94} & \underline{0.70} & \textbf{0.53} & \underline{0.96} & \underline{0.36} & \underline{0.45} & \underline{0.47} \\
			\hline
			(8,8,4)\cmark & \underline{0.71} & 0.56 & \textbf{0.71} & \underline{0.34} & \textbf{0.34} & 1.38 & 0.74 & \underline{0.64} & 1.34 & \underline{0.95} & \underline{0.70} & \underline{0.54} & 1.07 & \textbf{0.35} & \underline{0.45} & \underline{0.47} \\
			\hline
			Voxurf & 0.73 & 0.76 & \underline{0.72} & 0.67 & \textbf{0.34} & \underline{0.95} & \textbf{0.62} & 0.79 & 1.35 & 0.96 & 0.74 & 0.61 & 1.17 & \textbf{0.35} & \textbf{0.44} & 0.49 \\
			\hline
			Neus2 & 0.80 & \underline{0.55} & 0.81 & 1.66 & 0.38 & \textbf{0.92} & 0.72 & 0.79 & \underline{1.31} & 1.07 & 0.80 & 0.61 & \textbf{0.89} & 0.46 & 0.52 & 0.58 \\
			\hline
			2DGS & 0.76 & \textbf{0.47} & 0.82 & \textbf{0.32} & \underline{0.36} & 1.06 & 0.89 & 0.81 & \textbf{1.30} & 1.23 & \textbf{0.66} & 0.65 & 1.34 & 0.42 & 0.66 & \textbf{0.46} \\
			\hline
		\end{tabular}
	\caption{DTU per-scene chamfer (mm)}
	\end{center}
	\scriptsize
	\begin{center}
		\begin{tabular}{|c||c||c|c|c|c|c|c|c|c|c|c|c|c|c|c|c|}
			\hline
			PSNR (db) & Mean & 24 & 37 & 40 & 55 & 63 & 65 & 69 & 83 & 97 & 105 & 106 & 110 & 114 & 118 & 122 \\
			\hline
			(4,4,4)\cmark & 37.03 & \underline{35.58} & 30.59 & \underline{35.59} & \underline{35.56} & 38.89 & 38.06 & 34.81 & 39.90 & 33.97 & \underline{38.84} & \underline{40.03} & \underline{36.23} & 34.90 & 40.87 & 41.63 \\
			\hline
			(8,8,4)\cmark & \textbf{37.74} & \textbf{36.73} & \textbf{31.53} & \textbf{36.31} & \textbf{36.74} & \underline{39.38} & \underline{38.93} & 35.05 & \underline{40.25} & \textbf{34.69} & \textbf{39.51} & \textbf{40.45} & \textbf{36.79} & \textbf{35.60} & \underline{41.59} & \textbf{42.49} \\
			\hline
			Voxurf & \underline{37.08} & 34.97 & \underline{30.70} & 33.82 & 35.02 & \textbf{39.45} & \textbf{39.22} & \underline{35.48} & \textbf{41.03} & \underline{34.35} & 38.78 & 39.43 & 35.36 & \underline{35.20} & \textbf{41.79} & \underline{41.69} \\
			\hline
			Neus2 & 36.00 & 34.57 & 29.82 & 34.30 & 34.64 & 37.93 & 36.87 & 33.79 & 38.95 & 32.79 & 38.13 & 38.37 & 35.14 & 34.37 & 39.93 & 40.32 \\
			\hline
			2DGS & 36.03 & 35.01 & 30.61 & 34.47 & 33.77 & 38.27 & 36.11 & \textbf{35.84} & 39.53 & 34.26 & 38.33 & 37.86 & 34.82 & 33.46 & 39.10 & 39.05 \\
			\hline
		\end{tabular}
	\caption{DTU per-scene PSNR}
	\end{center}
	\scriptsize
	\begin{center}
		\begin{tabular}{|c||c||c|c|c|c|c|c|c|c|c|c|c|c|c|c|c|}
			\hline
			Chamfer (mm) & Mean & 24 & 37 & 40 & 55 & 63 & 65 & 69 & 83 & 97 & 105 & 106 & 110 & 114 & 118 & 122 \\
			\hline
			(4,4,4)\xmark & 0.71 & 0.68 & 0.82 & \textbf{0.34} & \underline{0.35} & 1.20 & 0.76 & \textbf{0.59} & 1.34 & \underline{0.91} & 0.74 & 0.57 & \textbf{0.91} & 0.39 & 0.50 & \underline{0.50} \\
			\hline
			(8,8,4)\cmark & 0.71 & \textbf{0.56} & \textbf{0.71} & \textbf{0.34} & \textbf{0.34} & 1.38 & 0.74 & 0.64 & 1.34 & 0.95 & \underline{0.70} & 0.54 & 1.07 & \textbf{0.35} & 0.45 & \textbf{0.47} \\
			\hline
			(12,12,4)\cmark & 0.70 & \underline{0.58} & \underline{0.74} & \textbf{0.34} & \textbf{0.34} & 1.37 & 0.73 & 0.66 & \underline{1.32} & \textbf{0.87} & 0.71 & 0.54 & 0.93 & \textbf{0.35} & 0.45 & \textbf{0.47} \\
			\hline
			(4,4,1)\cmark & 0.85 & 0.64 & 0.80 & \underline{0.35} & \textbf{0.34} & 1.79 & 0.71 & 0.77 & \textbf{1.30} & 1.12 & \textbf{0.69} & \textbf{0.52} & 2.34 & 0.41 & \textbf{0.43} & \textbf{0.47} \\
			\hline
			(4,4,2)\cmark & 0.69 & 0.64 & \underline{0.74} & \textbf{0.34} & \textbf{0.34} & 1.09 & \textbf{0.69} & 0.67 & 1.33 & 1.01 & \textbf{0.69} & \underline{0.53} & 1.02 & 0.37 & \underline{0.44} & \textbf{0.47} \\
			\hline
			(4,4,3)\cmark & \textbf{0.67} & 0.64 & 0.75 & \textbf{0.34} & \textbf{0.34} & \underline{1.04} & \underline{0.70} & \underline{0.62} & 1.33 & 0.95 & \textbf{0.69} & \underline{0.53} & \underline{0.92} & \underline{0.36} & \underline{0.44} & \textbf{0.47} \\
			\hline
			(4,4,4)\cmark & \underline{0.68} & 0.65 & \underline{0.74} & \textbf{0.34} & \textbf{0.34} & \textbf{1.02} & 0.71 & \underline{0.62} & 1.34 & 0.94 & \underline{0.70} & \underline{0.53} & 0.96 & \underline{0.36} & 0.45 & \textbf{0.47} \\
			\hline
		\end{tabular}
	\caption{Detailed Ablation Table. DTU per-scene chamfer (mm)}
	\end{center}
	\scriptsize
	\begin{center}
		\begin{tabular}{|c||c||c|c|c|c|c|c|c|c|c|c|c|c|c|c|c|}
			\hline
			PSNR (db) & Mean & 24 & 37 & 40 & 55 & 63 & 65 & 69 & 83 & 97 & 105 & 106 & 110 & 114 & 118 & 122 \\
			\hline
			(4,4,4)\xmark & 36.18 & 34.73 & 29.81 & 34.87 & 34.52 & 38.23 & 36.38 & 34.52 & 39.35 & 33.17 & 38.34 & 38.80 & 36 & 34.03 & 39.62 & 40.38 \\
			\hline
			(8,8,4)\cmark & \underline{37.74} & \underline{36.73} & \underline{31.53} & \underline{36.31} & \underline{36.74} & \underline{39.38} & \underline{38.93} & \underline{35.05} & \underline{40.25} & \underline{34.69} & \underline{39.51} & \textbf{40.45} & \underline{36.79} & \underline{35.60} & \underline{41.59} & \underline{42.49} \\
			\hline
			(12,12,4)\cmark & \textbf{38.03} & \textbf{37.20} & \textbf{31.96} & \textbf{36.64} & \textbf{36.95} & \textbf{39.55} & \textbf{39.33} & \textbf{35.71} & \textbf{40.42} & \textbf{35.27} & \textbf{39.58} & \underline{40.32} & \textbf{36.97} & \textbf{35.89} & \textbf{41.97} & \textbf{42.75} \\
			\hline
			(4,4,1)\cmark & 35.92 & 35.13 & 29.81 & 34.97 & 34.66 & 36.40 & 36.77 & 32.90 & 38.86 & 32.29 & 38.03 & 39.43 & 34.36 & 34 & 40.36 & 40.92 \\
			\hline
			(4,4,2)\cmark & 36.44 & 35.25 & 29.97 & 35.31 & 35.27 & 37.66 & 37.32 & 33.62 & 39.48 & 32.98 & 38.53 & 39.71 & 35.49 & 34.39 & 40.53 & 41.11 \\
			\hline
			(4,4,3)\cmark & 36.76 & 35.36 & 30.27 & 35.40 & 35.46 & 38.46 & 37.66 & 34.40 & 39.71 & 33.59 & 38.67 & 39.80 & 35.95 & 34.67 & 40.69 & 41.32 \\
			\hline
			(4,4,4)\cmark & 37.03 & 35.58 & 30.59 & 35.59 & 35.56 & 38.89 & 38.06 & 34.81 & 39.90 & 33.97 & 38.84 & 40.03 & 36.23 & 34.90 & 40.87 & 41.63 \\
			\hline
		\end{tabular}
	\caption{Detailed Ablation Table. DTU per-scene PSNR}
	\end{center}
\end{table*}


\begin{table*}[htb]
	\begin{center}
	\small
		\begin{tabular}{|c||c||c|c|c|c|c|c|c|c|}
			\hline
			Chamfer & Mean & dog & bear & clock & durian & man & sculpture & stone & jade \\
			\hline
			(4,4,4)\cmark & \underline{2.36} & 2.51 & 2.27 & 1.90 & 3.63 & \textbf{1.79} & \textbf{1.61} & \textbf{1.30} & \underline{3.88} \\
			\hline
			(8,8,4)\cmark & \textbf{2.21} & \textbf{2.24} & \textbf{2.03} & \textbf{1.69} & \underline{3.26} & \underline{1.81} & \textbf{1.61} & \underline{1.36} & \textbf{3.71} \\
			\hline
			Voxurf & 2.64 & \underline{2.28} & \underline{2.20} & \underline{1.88} & \textbf{2.98} & 2.11 & \underline{1.75} & 3.96 & 3.99 \\
			\hline
			Neus2 & 2.93 & 2.78 & 2.71 & 2.63 & 4.23 & 2.25 & 2.50 & 1.91 & 4.43 \\
			\hline
		\end{tabular}
	\caption{BlendedMVS per-scene chamfer}
	\label{tab:bmvs_chamfer}
	\end{center}
	\begin{center}
	\small
		\begin{tabular}{|c||c||c|c|c|c|c|c|c|c|}
			\hline
			PSNR & Mean & dog & bear & clock & durian & man & sculpture & stone & jade \\
			\hline
			(4,4,4)\cmark & \underline{35.19} & \underline{35.49} & 30.55 & \underline{34.68} & \underline{31.28} & 42.94 & 40.80 & 30.84 & 34.92 \\
			\hline
			(8,8,4)\cmark & \textbf{35.89} & \textbf{36.30} & \textbf{30.96} & \textbf{35.42} & \textbf{31.65} & \textbf{43.72} & \textbf{41.34} & \underline{31.01} & \textbf{36.69} \\
			\hline
			Voxurf & 35.11 & 35.24 & \underline{30.88} & 34.49 & 29.69 & \underline{43.35} & \underline{41.06} & 30.23 & \underline{35.93} \\
			\hline
			Neus2 & 33.62 & 34.56 & 29.99 & 31.04 & 29.21 & 40.88 & 39.10 & \textbf{31.36} & 32.79 \\
			\hline
		\end{tabular}
	\caption{BlendedMVS per-scene PSNR}
	\end{center}
	\begin{center}
	\small
		\begin{tabular}{|c||c||c|c|c|c|c|c|c|c|}
			\hline
			Chamfer & Mean & dog & bear & clock & durian & man & sculpture & stone & jade \\
			\hline
			(4,4,4)\xmark & 2.47 & \underline{2.18} & 2.71 & 2.05 & 3.81 & 1.97 & 1.75 & 1.36 & 3.95 \\
			\hline
			(8,8,4)\cmark & \textbf{2.21} & 2.24 & \underline{2.03} & \textbf{1.69} & \textbf{3.26} & 1.81 & \underline{1.61} & 1.36 & \textbf{3.71} \\
			\hline
			(12,12,4)\cmark & \underline{2.31} & \textbf{2.10} & 2.31 & \underline{1.86} & \underline{3.58} & 1.96 & \textbf{1.59} & 1.34 & \underline{3.74} \\
			\hline
			(4,4,1)\cmark & 2.58 & 3.27 & \textbf{2.02} & 2.44 & 3.88 & 1.92 & 1.88 & 1.33 & 3.93 \\
			\hline
			(4,4,2)\cmark & 2.35 & 2.59 & \underline{2.03} & 2.04 & 3.66 & \underline{1.80} & 1.62 & \textbf{1.29} & 3.75 \\
			\hline
			(4,4,3)\cmark & 2.36 & 2.56 & 2.35 & 1.94 & 3.59 & \textbf{1.79} & 1.64 & 1.31 & \underline{3.74} \\
			\hline
			(4,4,4)\cmark & 2.36 & 2.51 & 2.27 & 1.90 & 3.63 & \textbf{1.79} & \underline{1.61} & \underline{1.30} & 3.88 \\
			\hline
		\end{tabular}
	\end{center}
	\caption{Detailed Ablation Table. BlendedMVS per-scene chamfer}
	\begin{center}
	\small
		\begin{tabular}{|c||c||c|c|c|c|c|c|c|c|}
			\hline
			(4,4,4)\xmark & 34.76 & 35.71 & 30.35 & 33.76 & 31 & 42.65 & 40.08 & 30.84 & 33.72 \\
			\hline
			(8,8,4)\cmark & \underline{35.89} & \underline{36.30} & \underline{30.96} & \underline{35.42} & \underline{31.65} & \underline{43.72} & \underline{41.34} & \underline{31.01} & \underline{36.69} \\
			\hline
			(12,12,4)\cmark & \textbf{36.17} & \textbf{36.50} & \textbf{30.97} & \textbf{36.55} & \textbf{32.12} & \textbf{43.75} & \textbf{41.41} & \textbf{31.04} & \textbf{37.00} \\
			\hline
			(4,4,1)\cmark & 34.44 & 34.48 & 30.34 & 32.81 & 31.12 & 42.44 & 39.89 & 30.75 & 33.72 \\
			\hline
			(4,4,2)\cmark & 34.86 & 35 & 30.53 & 33.85 & 31.38 & 42.71 & 40.40 & 30.82 & 34.19 \\
			\hline
			(4,4,3)\cmark & 35.05 & 35.18 & 30.64 & 34.34 & 31.36 & 42.90 & 40.69 & 30.82 & 34.48 \\
			\hline
			(4,4,4)\cmark & 35.19 & 35.49 & 30.55 & 34.68 & 31.28 & 42.94 & 40.80 & 30.84 & 34.92 \\
			\hline
		\end{tabular}
	\end{center}
	\caption{Detailed Ablation Table. BlendedMVS per-scene PSNR}
	\label{tab:bmvs_ablations_psnr}
\end{table*}

\end{document}